\documentclass[10pt,twocolumn,letterpaper]{article}

\usepackage[pagenumbers]{cvpr} 

\definecolor{cvprblue}{rgb}{0.21,0.49,0.74}
\usepackage[pagebackref,breaklinks,colorlinks,allcolors=cvprblue]{hyperref}

\usepackage{url}            
\usepackage{booktabs}       
\usepackage{amsfonts}       
\usepackage{nicefrac}       
\usepackage{microtype}      
\usepackage{xcolor}         
\usepackage{graphicx,wrapfig,amsmath}
\usepackage[noend]{algorithm2e}
\usepackage[noend]{algpseudocode}

\usepackage{pifont}
\newcommand{\cmark}{\ding{51}}%
\newcommand{\xmark}{\ding{55}}%

\usepackage{xspace}
\makeatletter
\DeclareRobustCommand\onedot{\futurelet\@let@token\@onedot}
\def\@onedot{\ifx\@let@token.\else.\null\fi\xspace}
\def\eg{\emph{e.g}\onedot} 
\def\ie{\emph{i.e}\onedot}

\def\wrt{w.r.t\onedot} 

\usepackage{makecell}

\usepackage{amssymb}
\usepackage{color,colortbl}

\usepackage{multirow}
\usepackage{pifont}
\usepackage{gensymb}
\usepackage{enumitem}

\definecolor{lightgreen}{RGB}{100,220,100}
\definecolor{darkgreen}{RGB}{30,150,30}
\definecolor{darkblue}{RGB}{0,0,127}
\definecolor{darkyellow}{RGB}{171,133,0}
\definecolor{darkred}{RGB}{180,20,20}
\definecolor{darkmagenta}{RGB}{127,0,127}
\definecolor{darkcyan}{RGB}{0,127,127}

\definecolor{purple}{HTML}{9900ff}
\definecolor{darkpink}{HTML}{ff00ff}
\definecolor{maroon}{HTML}{980000}

\definecolor{lightred}{RGB}{220,20,20}

\newcommand{\darkgreen}[1]{\textcolor{darkgreen}{#1}}

\usepackage{stmaryrd}

\newcommand{\cvprcolor}[1]{\textcolor{cvprblue}{#1}}

\newcommand{\revision}[1]{#1}

\usepackage{lipsum}
\usepackage{bbm}

\definecolor{darkpastelgreen}{rgb}{0.01, 0.75, 0.24}

\def\paperID{1495} 
\def\confName{CVPR}
\def\confYear{2025}

\title{Interpretable Generative Models through Post-hoc Concept Bottlenecks}

\author{Akshay Kulkarni, Ge Yan\thanks{Equal contribution}, Chung-En Sun\footnotemark[1], Tuomas Oikarinen, and Tsui-Wei Weng\\
University of California San Diego\\
{\tt\small \{a2kulkarni, lweng\}@ucsd.edu}
}

\begin{document}

\maketitle

\begin{abstract}
Concept bottleneck models (CBM) aim to produce inherently interpretable models that rely on human-understandable concepts for their predictions. However, existing approaches to design interpretable generative models based on CBMs are not yet efficient and scalable, as they require expensive generative model training from scratch as well as real images with labor-intensive concept supervision. To address these challenges, we present two novel and low-cost methods to build interpretable generative models through post-hoc techniques and we name our approaches: concept-bottleneck autoencoder (CB-AE) and concept controller (CC). Our proposed approaches enable efficient and scalable training without the need of real data and require only minimal to no concept supervision. Additionally, our methods generalize across modern generative model families including generative adversarial networks and diffusion models. We demonstrate the superior interpretability and steerability of our methods on numerous standard datasets like CelebA, CelebA-HQ, and CUB with large improvements (average $\sim$25\%) over the prior work, while being 4-15$\times$ faster to train. Finally, a large-scale user study is performed to validate the interpretability and steerability of our methods.
\vspace{-5mm}
\end{abstract}

\footnotetext[1]{Code: \href{https://github.com/Trustworthy-ML-Lab/posthoc-generative-cbm}{github.com/Trustworthy-ML-Lab/posthoc-generative-cbm}}

\begin{figure}[t]
  \centering
  \includegraphics[width=\linewidth]{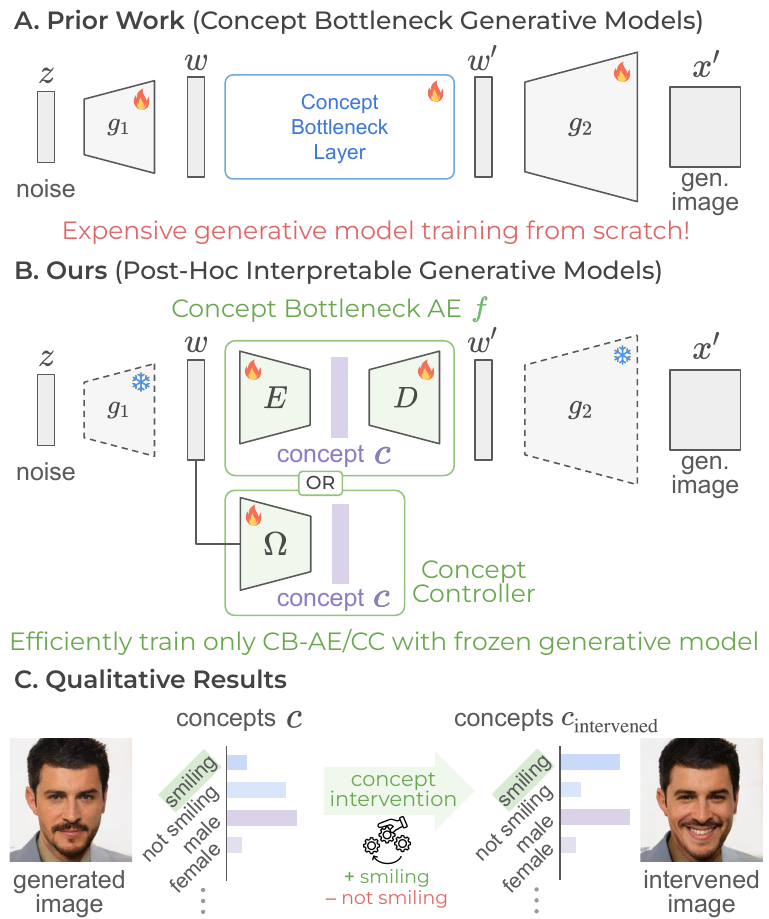}
  \vspace{-7mm}
  \caption{
  \textbf{A.} Prior work on interpretable generative models requires expensive generative model training from scratch. \textbf{B.} Our CB-AE and CC can be trained efficiently for post-hoc interpretability in a pretrained, frozen generative model $g_2\!\circ\! g_1$. \textbf{C.} Example concept intervention with CB-AE and corresponding concept vectors.
  }
  \vspace{-4mm}
  \label{fig:post_hoc_setup}
\end{figure}

\section{Introduction}
\vspace{-1.5mm}

Deep generative models 
\cite{rombach2022high,saharia2022photorealistic,ramesh2021zero} have become increasingly powerful and widely used in many high-stakes domains and applications, including realistic data generation \cite{he2023synthetic}, simulating hypothetical scenarios or environments \cite{katara2024gensim}, and scientific discovery \cite{anstine2023generative}. It is therefore important to ensure that the generation process is interpretable, which will allow us to understand and audit the generation, and further mitigate potential biases and harms (\eg content moderation). 

Unfortunately, most of the advances in deep learning utilize complex, black-box neural network architectures that are difficult to interpret and understand. This leads to user mistrust of model predictions due to the absence of explanations. To address this, there has been work on developing inherently interpretable deep vision models using concept bottlenecks \cite{koh2020concept,yuksekgonul2023post,oikarinen2023label,srivastava2024vlg,yan2023learning,yang2023language}. These approaches train a concept bottleneck layer after the feature extractor (backbone) to embed a set of human-understandable concepts, followed by an interpretable sparse linear layer for the final classification based on the concept prediction. However, current development of CBMs is primarily focused on classification tasks, and only one prior work, CBGM \cite{ismail2024concept}, has extended it to image generation, indicating this area is under-explored. 

\begin{table*}[t]
\caption{Comparison of our CB-AE and CC with prior work CBGM \cite{ismail2024concept}
(\textcolor{darkgreen}{green} indicates desirable properties).
CC trades-off 
inherent interpretability
for better steerability, image quality, and faster training. We could not reproduce CBGM \cite{ismail2024concept} to evaluate concept accuracy.
}
\label{tab:char_comp}
\vspace{-3mm}
\setlength{\tabcolsep}{3mm}
\resizebox{\linewidth}{!}{
\begin{tabular}{lcccclcll}
\toprule
\multirow{2}{*}{Method} & \multicolumn{1}{c}{\multirow{2}{*}{\begin{tabular}[c]{@{}c@{}}Post-hoc\\ training\end{tabular}}} & \multicolumn{1}{c}{\multirow{2}{*}{\begin{tabular}[c]{@{}c@{}}Training without concept-\\ labeled real images\end{tabular}}} & \multicolumn{1}{c}{\multirow{2}{*}{\begin{tabular}[c]{@{}c@{}}Inherently\\ interpretable model\end{tabular}}} & \multicolumn{1}{c}{\multirow{2}{*}{\begin{tabular}[c]{@{}c@{}}Concept\\ Acc. (\%)\end{tabular}}} & \multicolumn{1}{c}{\multirow{2}{*}{\begin{tabular}[c]{@{}c@{}}Steerability\\ (\%) ($\uparrow$)\end{tabular}}} & \multicolumn{1}{c}{\multirow{2}{*}{\begin{tabular}[c]{@{}c@{}}FID\\ ($\downarrow$)\end{tabular}}} & \multicolumn{2}{c}{Train time (V100-hrs) ($\downarrow$)} \\ \cmidrule(lr){8-9}
& & & & & & & \multicolumn{1}{c}{\small{StyleGAN2}} & \multicolumn{1}{c}{\small{DDPM}} \\ 
\midrule
CBGM \cite{ismail2024concept} 
  & \textcolor{lightred}{\xmark} 
  & \textcolor{lightred}{\xmark} 
  & \textcolor{darkgreen}{\cmark} 
  & - 
  & 25.60 
  & 9.10 
  & 50 
  & 240 \\
\rowcolor{gray!10}CB-AE (\textit{Ours}) 
  & \textcolor{darkgreen}{\cmark} 
  & \textcolor{darkgreen}{\cmark} 
  & \textcolor{darkgreen}{\cmark} 
  & 86.56 
  & 47.34 \small{\textcolor{darkgreen}{(+21.74)}} 
  & 9.52 
  & 14 \small{\textcolor{darkgreen}{(3.5$\times$ faster)}} 
  & 29.5 \small{\textcolor{darkgreen}{(8.1$\times$ faster)}} \\
\rowcolor{gray!10}CC (\textit{Ours}) 
  & \textcolor{darkgreen}{\cmark} 
  & \textcolor{darkgreen}{\cmark} 
  & \textcolor{lightred}{\xmark} 
  & \textbf{87.65} 
  & \textbf{51.14} \small{\textcolor{darkgreen}{(+25.54)}} 
  & \textbf{7.65} 
  & \textbf{6} \small{\textcolor{darkgreen}{(8.3$\times$ faster)}} 
  & \textbf{8.3} \small{\textcolor{darkgreen}{(28.9$\times$ faster)}} \\
\bottomrule
\end{tabular}
}
\vspace{-3mm}
\end{table*}

\revision{The key idea of the seminal work} CBGM~\cite{ismail2024concept} is to represent concepts as learnable embeddings~\cite{zarlenga2022concept} at an intermediate location in the generative model,
and combine the embeddings to compute the generative model latent. However, the CBGM-based generative model has to be trained from scratch using concept-labeled real images, which could be difficult to scale and computationally intensive (\eg 240 V100-hours for DDPM-256$\times$256 \cite{ho2020denoising}) as shown in Fig.\ \ref{fig:post_hoc_setup}\cvprcolor{A}. To address these limitations, in this work, 
we develop an \textit{efficient} and \textit{scalable} post-hoc concept bottleneck to \textit{transform} any pretrained generative model into an interpretable model.  
In contrast to CBGM~\cite{ismail2024concept}, our approach works well by only training a few layers with minimal concept supervision.

Specifically, we propose a novel concept-bottleneck autoencoder (CB-AE) $f$ that can be inserted into the intermediate layers of a pretrained generative model $g = g_2\circ g_1$ as shown in Fig.\ \ref{fig:post_hoc_setup}\cvprcolor{B} (top). The CB-AE input and output are the generator latent $w$ and its reconstruction $w'$ respectively, and the overall generative model now becomes $g_2 \circ f \circ g_1$. The CB-AE latent space is the concept space $c$ which is used to reconstruct the generator latent space. In our framework, given a pretrained generative model $g\!=\!g_2\circ g_1$, only the encoder $E$ and decoder $D$ in the CB-AE $f=D\circ E$ need to be trained, while $g_1, g_2$ are frozen pretrained weights, \ie CB-AE uses post-hoc training. The benefit of the proposed CB-AE is that it allows us to debug the model easily with concept-level control by modifying the CB-AE concept latent during image generation (Fig.\ \ref{fig:post_hoc_setup}\cvprcolor{C}). Further, our training requires only concept pseudo-labels achievable with minimal concept supervision (\eg zero-shot CLIP classifier \cite{radford2021learning}).

We also propose novel optimization-based concept interventions to achieve higher success rate and intervention quality. Based on the CB-AE, we propose an even more efficient post-hoc concept controller (CC) method (Fig.\ \ref{fig:post_hoc_setup}\cvprcolor{B}, bottom) with simplified training, that can provide concept predictions and leverage 
optimization-based concept interventions. 
\revision{Note that while CB-AE is part of the new interpretable generative model, CC is a post-hoc control method that is not part of the generative model.}
Finally, we evaluate our CB-AE and CC on various generative models, including GANs and diffusion models, for standard datasets including CelebA, CelebA-HQ, and CUB. We show that CB-AE and CC significantly outperform prior work CBGM \cite{ismail2024concept} \wrt steerability (or intervention success rate) on CelebA (average \darkgreen{+23\%}) while also having \darkgreen{4-15$\times$ faster} training (Table \ref{tab:char_comp}).

\noindent
Our contributions can be summarized as follows:
\begin{itemize}
    \itemsep0em
    \item We are the first to propose a \textit{post-hoc} concept bottleneck autoencoder (CB-AE) for interpretable generative models. CB-AE can be trained efficiently with a frozen pretrained generative model, without real concept-labeled images.
    \item We also propose a novel and efficient optimization-based concept intervention method with improved steerability (avg.\ \darkgreen{+19\%}) and higher image quality (avg.\ \darkgreen{32\%} better).
    \item We validate the effectiveness of our methods for GANs and diffusion models (avg.\ \darkgreen{+31\%} and \darkgreen{+28\%} steerability \wrt prior state-of-the-art) across varying image resolutions, while being \darkgreen{4-15$\times$} faster to train on average.
\end{itemize}

\begin{figure*}[t]
    \centering
    \includegraphics[width=0.9\linewidth]{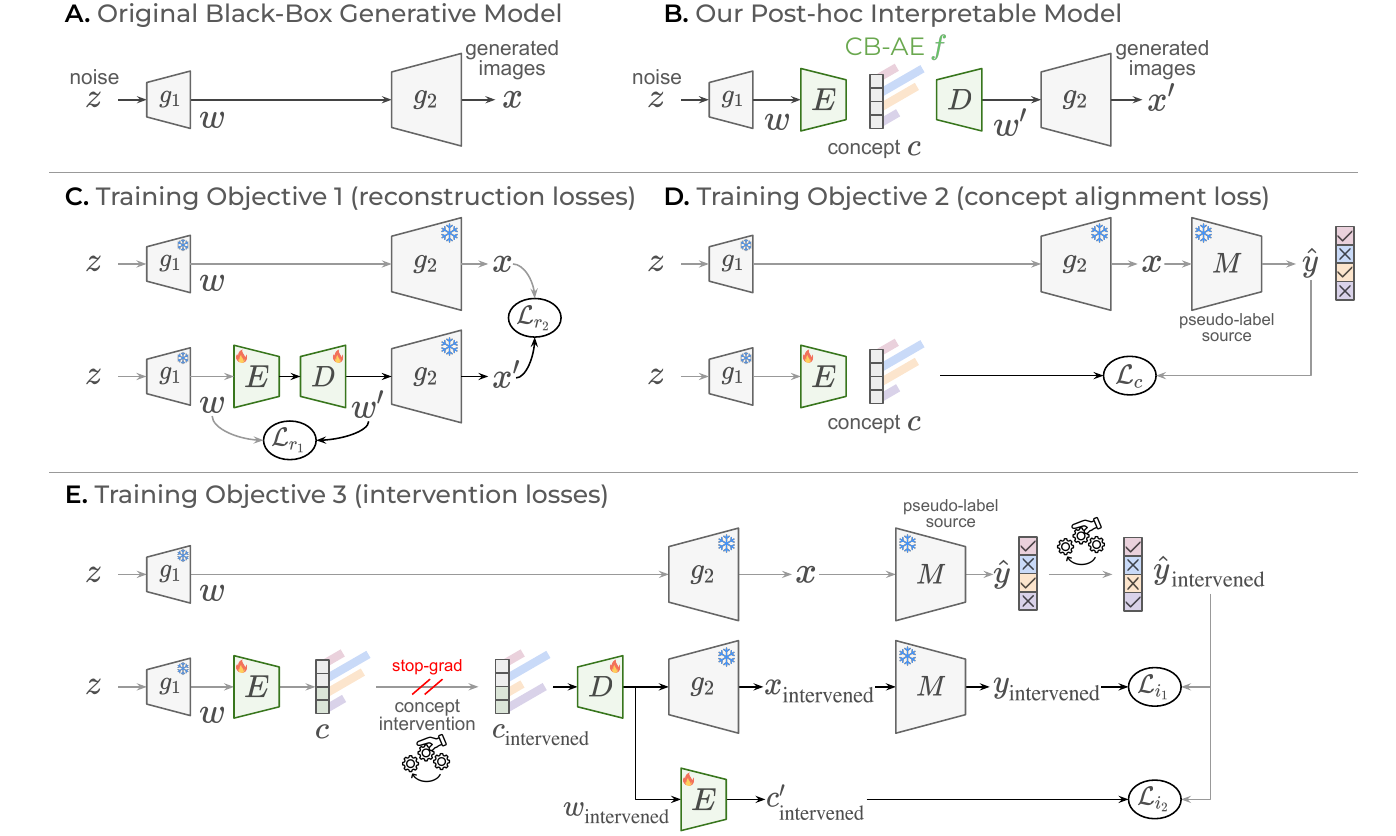}
    \vspace{-3mm}
    \caption{\textbf{Post-hoc CB-AE training} for reconstruction, concept alignment, and intervention with a frozen pretrained generator $g_2\!\circ\! g_1$. Note that $\mathcal{L}_c$, $\mathcal{L}_i$ indicate cross-entropy loss and $\mathcal{L}_r$ indicates mean-squared-error loss. The darker lines indicate gradient flow during training.
    }
    \label{fig:stylegan_approach}
\end{figure*}

\section{Related Work}

\noindent
\textbf{Concept Bottleneck Models.} 
Early work on CBMs \cite{koh2020concept, marconato2022glancenets} relied on concept-labeled images to train a concept bottleneck layer with each neuron as a human-understandable concept, followed by a linear layer based on the concepts for the final classification. 
Post-hoc CBMs \cite{yuksekgonul2023post} extended this idea to convert a pretrained backbone into a CBM. 
More recent works like LF-CBM \cite{oikarinen2023label}, LM4CV \cite{yan2023learning}, LaBo \cite{yang2023language}, and VLG-CBM \cite{srivastava2024vlg} use interpretability tools \cite{oikarinen2023clipdissect}, vision-language models like CLIP \cite{radford2021learning}, large language models, or open-set object detection \cite{liu2024grounding} to eliminate the need for expensive concept-labeled data.
Independently from our work, \cite{laguna2024beyond} proposed a concept-based intervention without a concept bottleneck, similar to our CC, but for classification.
All these works are specifically designed for classification, while in this paper, our focus is on the image generation task.

\noindent
\textbf{Interpretability for Generative Models.} 
A line of work \cite{higgins2017betavae, chen2018isolating, ding2020guided} on learning disentangled concepts in variational autoencoders enables controllable generation, but they train from scratch and are not applicable to other generative models like GANs.
Other works focus on identifying and manipulating structural rules or concepts in GANs \cite{bau2019gandissect,bau2020rewriting} and large language models \cite{mitchell2022fast,mitchell2022memory,meng2022locating,sun2025cbllm} by editing the model weights. 
In contrast, we focus on training inherently interpretable models, \revision{and the closest prior work is the recent} CBGM \cite{ismail2024concept}, which also aims to build interpretable generative models, but requires expensive concept labels and training from scratch, limiting its scalability. In contrast, our proposed CB-AE can be trained efficiently with a frozen pretrained generative model with minimal concept supervision.

\noindent
\textbf{Image Editing in Generative Models.} 
Some works focus on conditional generation \cite{rombach2022high,saharia2022photorealistic} and image editing in generative models by modifying model weights \cite{gandikota2024sliders, rishubh2024precisecontrol}. In contrast, our work focuses on developing inherently interpretable generative models by 
introducing additional concept bottleneck layers, where the capability of editing or intervention is naturally a by-product of the interpretability.

\section{Proposed Methods}

We propose a novel and low-cost concept bottleneck autoencoder (CB-AE) method in Sec.\ \ref{subsec:cbae} to incorporate post-hoc interpretability in pretrained generative models. 
In Sec.\ \ref{subsec:opt-int}, we present an optimization-based intervention method for concept-based steerability in the generative model. Finally, based on our insights from CB-AE and optimization-based interventions, we propose an even lower-cost post-hoc \revision{control} method, concept controller (CC) in Sec.\ \ref{subsec:cc}.

\noindent
\textbf{Preliminaries.}
Consider a generative model $g:\mathcal{Z}\to\mathcal{X}$ that maps from random noise $z \in \mathcal{Z}$ to an image $x\in\mathcal{X}$. 
To make the generative model $g$ become inherently interpretable, the goal of CBMs is to insert and train a concept bottleneck (say) $f$ at an intermediate location in $g$. Let $g=g_2 \circ g_1$, \ie $g$ can be divided into two parts, $g_1$ and $g_2$ (\eg for a DCGAN \cite{radford2016dcgan}, $g_1$ is the first 2 layers of $g$ and $g_2$ is the remaining layers). 
The input to the concept bottleneck $f$ will be the output of $g_1$ (Fig.\ \ref{fig:post_hoc_setup}). At the bottleneck of $f$, we obtain the concept prediction $c\in \mathcal{C}$ where $\mathcal{C}$ is the set of pre-defined concepts. In our setup, the concept vector $c$ has two logits for each binary concept $c_i$ (\eg ``smiling" would have two logits, for smiling and not smiling) and $N$ logits for each categorical concept $c_j$ with $N$ classes (\eg blonde/black/white/gray hair color would have four logits). For example, suppose we have one binary concept $c_i$ and one categorical concept $c_j$, then we define $c \!=\! [c_i^+, c_i^-, c_j^{(1)}, c_j^{(2)}, \ldots, c_j^{(N)}]^\top$ where $c_i\!=\![c_i^+, c_i^-]^\top\!\in\! \mathbb{R}^2$ and $c_j\!=\![c_j^{(1)}, c_j^{(2)}, \ldots, c_j^{(N)}]^\top\!\in \!\mathbb{R}^N$.

\subsection{Post-hoc Concept Bottleneck Autoencoder}
\label{subsec:cbae}

We propose a concept bottleneck autoencoder (CB-AE), $f=D \circ E$ (see Fig.\ \ref{fig:stylegan_approach}\cvprcolor{B}). The latent space of the CB-AE encoder $E$ is the concept prediction $c = E(g_1(z))$. 
Apart from the predefined concepts, $c$ also contains an unsupervised concept embedding, learned using autoencoder reconstruction and intervention objectives, to encode other concepts absent from the predefined set (similar to CBGM \cite{ismail2024concept}).
The decoder $D$ reconstructs the features from $g_1$ based on the concept prediction $c$, outputting $w' = D(c)$. Considering the original output of $g_1$ to be $w=g_1(z)$, we can generate the original image $x = g_2(w)$ as well as a reconstructed image $x' \!=\! g_2(w')\!=\!g_2(D\circ E(w))$ generated using the CB-AE.

\noindent
\textbf{Training.}
There are 3 goals for the CB-AE. First, the generator's performance should be preserved even if the CB-AE output $w'$ is used instead of $w$. Second, CB-AE should provide interpretability for the generated images $x'$ through the corresponding concepts $c$. Lastly, CB-AE should allow accurate steering on image generation via concept interventions. Based on these 3 goals, we formulate  \textbf{Objective 1-3} for CB-AE training below and illustrated in Fig.\ \ref{fig:stylegan_approach}\cvprcolor{C}-\cvprcolor{E}.

\noindent
\textbf{Objective 1 (reconstruction losses $\boldsymbol{\mathcal{L}_{r_1}, \mathcal{L}_{r_2}}$).} At each training iteration, we sample a latent $w$ by passing uniform noise $z$ to $g_1$. Then, the latent $w$ is passed through $g_2$ to obtain a generated image $x=g_2(w)$ from the original model $g\!=\!g_2\circ g_1$, without using our CB-AE. 
We also reconstruct $w'=D \circ E(w)$ using our CB-AE and the same latent $w$, and obtain another generated image $x'=g_2(w')$. Since our first goal is to preserve the generator's performance when using the CB-AE, we apply reconstruction losses (mean-squared error loss $\mathcal{L}_r$) between the latents $w, w'$ and between the generated images $x, x'$, as shown in Fig.\ \ref{fig:stylegan_approach}\cvprcolor{C}:
\begin{equation}
    \min_{E, D} [\mathcal{L}_{r_1}(w, w') + \mathcal{L}_{r_2}(x, x')],
    \label{eqn:recon_loss}
\end{equation}
where $w'=D\circ E(w)$ and $x'=g_2(w')=g_2\circ D\circ E(w)$ and only the CB-AE parameters (\ie $E, D$) are trainable.

\noindent
\textbf{Objective 2 (concept alignment loss $\boldsymbol{\mathcal{L}_c}$).}
To ensure interpretability, we first obtain a concept pseudo-label $\hat{y}=M(x)$ for a generated image $x$ from a pseudo-label source $M$ (Fig.\ \ref{fig:stylegan_approach}\cvprcolor{D}). The source $M$ can be either an off-the-shelf supervised model or a zero-shot prediction pipeline from (say) CLIP with concept names $\mathcal{C}$ as the text inputs.
With this approach, we avoid the requirement of any real images for training as well as the requirement of concept labels, unlike CBGM \cite{ismail2024concept}. Since our second goal is to provide interpretability for the generated images through the concepts $c$, we apply a cross-entropy loss $\mathcal{L}_c$ between the CB-AE encoder output $c=E(w)$ and the concept pseudo-label $\hat{y}=M(x)$:
\begin{equation}
    \min_{E} [\mathcal{L}_c(\hat{y}, c)].
    \label{eqn:concept_loss}
\end{equation}
The losses in Eq.\ \eqref{eqn:recon_loss}, \eqref{eqn:concept_loss} are simultaneously optimized to learn only the CB-AE parameters $E, D$.

\noindent
\textbf{Objective 3 (intervention losses $\boldsymbol{\mathcal{L}_{i_1}, \mathcal{L}_{i_2}}$).}
Interventions are an important feature of concept-bottleneck models \cite{koh2020concept}, allowing users to control the model output by modifying the concepts $c$. However, for our CB-AE decoder $D$, reconstruction and concept alignment losses in Objective 1 and 2 do not provide guidance on how the reconstructed latent $w'$ should change when concepts $c$ are manually modified. Hence, for steerability, we design Objective 3 (Fig.\ \ref{fig:stylegan_approach}\cvprcolor{E}) that encourages the CB-AE decoder $D$ to produce an appropriately changed and realistic latent $w'$ when the concepts in $c$ are modified. We first describe \textbf{(a)} how interventions are performed in our CB-AE, followed by designing \textbf{(b)} an intervened concept alignment loss $\mathcal{L}_{i_1}$ and \textbf{(c)} a cyclic intervened concept loss $\mathcal{L}_{i_2}$ to encourage steerability, \ie intervention success.

\noindent
\textbf{\underline{a) Intervening concepts.}}
At each training iteration, we choose a random logit for a random concept to intervene, and modify only the chosen concept based on the chosen logit to get an intervened concept vector $c_\text{intervened}$ (Fig.\ \ref{fig:stylegan_approach}\cvprcolor{E}). 
For example, for a desired binary concept $i\in \mathcal{C}$, the new concept vector $c_\text{intervened}$ is computed by swapping the two logits, \ie $c_\text{intervened}=[\ldots, c_i^-, c_i^+, \ldots]$ from $c=[\ldots, c_i^+, c_i^-, \ldots]$.
The same can be done for a categorical concept $i\in \mathcal{C}$ as well by swapping the desired logits (say) $c_i^{(k)}$ with the highest logits $c_i^{(\ell)}$ where $\ell = \arg\max_j c_i^{(j)}$. Concretely, $c = [\ldots, c_i^{(1)}, \ldots, c_i^{(\ell)}, \ldots, c_i^{(k)}, \ldots, c_i^{(N)}, \ldots]$ is modified to $c_\text{intervened} = [\ldots, c_i^{(1)}, \ldots, c_i^{(k)}, \ldots c_i^{(\ell)}, \ldots, c_i^{(N)}, \ldots]$. 
We can also intervene on multiple concepts simultaneously.

\noindent
\textbf{\underline{b) Designing $\boldsymbol{\mathcal{L}_{i_1}}$.}}
Using the intervened concepts $c_\text{intervened}$, we reconstruct an intervened latent $w_\text{intervened}=D(c_\text{intervened})$ and obtain an intervened generated image $x_\text{intervened} = g_2(w_\text{intervened})$. Using the pseudo-label source, we obtain an intervened concept prediction $y_\text{intervened}=M(x_\text{intervened})$. Since we already have the concept pseudo-label $\hat{y}=M(x)$ for the original image $x$, we modify $\hat{y}$ to $\hat{y}_\text{intervened}$ by changing only the \revision{earlier} chosen concept \revision{$i\in \mathcal{C}$} in $\hat{y}$ to the chosen value \revision{(based on earlier chosen logit)} as shown in Fig.\ \ref{fig:stylegan_approach}\cvprcolor{E}.
In other words, $\hat{y}_\text{intervened}$ is the intervened concept pseudo-label. Now, to align the concepts in the intervened image $x_\text{intervened}$ with the predicted concepts from $M$, we use a cross-entropy loss $\mathcal{L}_{i_1}$ between $\hat{y}_\text{intervened}$ and $y_\text{intervened}$:
\begin{align}
    \min_{E, D} [\mathcal{L}_{i_1}\!(\hat{y}_\text{intervened}, y_\text{intervened})].
    \label{eqn:int_loss1}
\end{align}

\noindent
\textbf{\underline{c) Designing $\boldsymbol{\mathcal{L}_{i_2}}$.}}
For cyclic consistency, we pass the intervened latent $w_\text{intervened}$ through the CB-AE encoder $E$ to obtain a concept prediction $c'_\text{intervened}$ (Fig.\ \ref{fig:stylegan_approach}\cvprcolor{E}) and apply a cross-entropy loss $\mathcal{L}_{i_2}$ \wrt $\hat{y}_\text{intervened}$ to align the encoder's prediction with that of the pseudo-label source $M$: 
\begin{equation}
    \min_{E, D} [\mathcal{L}_{i_2}(\hat{y}_\text{intervened}, c'_\text{intervened})].
    \label{eqn:int_loss2}
\end{equation}
We use $\mathcal{L}_{i_1}, \mathcal{L}_{i_2}$ instead of $\mathcal{L}_c$ for cross-entropy loss to differentiate the intervention losses from the concept loss in Objective 2, and only CB-AE parameters $E, D$ are trainable. Finally, recall that the concept vector $c$ contains \revision{an unsupervised concept embedding.} Objective 3 implicitly encourages \revision{this embedding} to not encode known concepts, since \revision{the unsupervised embedding is} not modified during the intervention from $c$ to $c_\text{intervened}$.

\noindent
\textbf{Test-Time Intervention.}
Similar to training-time interventions, we can perform test-time interventions by modifying the value of any chosen concept (by swapping the logits) in the predicted concept vector $c$ to $c_\text{intervened}$. The intervened image $x_\text{intervened}=g_2(w_\text{intervened})$ can be obtained where $w_\text{intervened} = D(c_\text{intervened})$.
Note that swapping the logits ensures that the range of values in $c_\text{intervened}$ are similar to $c$, and also avoids the requirement of any estimation of how much to change the desired concept's logit. This makes it more accessible to users as they need not worry about the actual values being changed during an intervention.

\subsection{Optimization-based interventions}
\label{subsec:opt-int}

For an alternative intervention method, we draw inspiration from adversarial attacks \cite{goodfellow2015explaining} to perform test-time interventions using gradient-based optimization. Specifically, we use the iterative randomized fast gradient sign method (I-RFGSM) \cite{wong2020fast} on the CB-AE encoder prediction.

Consider a generated image $x=g_2(w)$ with concept prediction $c=E(w)$. To intervene in the generation process to obtain modified concepts $c^*$, we solve the following objective using gradient ascent,
\begin{align}
    w^* = w + \mathop{\arg\max}_{\delta \in \Delta} [-\mathcal{L}_c(E(w+\delta), c^*)]
\end{align}
where $\Delta=\{\delta: \Vert\delta\Vert_\infty \leq \epsilon\}$ is the $\ell_\infty$-norm bound on $\delta$ with a hyperparameter $\epsilon>0$. Intuitively, we optimize a small perturbation $\delta$ such that $w^*=w+\delta$ leads to the desired concepts $c^*$. Then, the generated image $x^*=g_2(w^*)$ is very similar to $x$ but contains concepts $c^*$.

\begin{figure*}[t]
  \centering

  \begin{minipage}{\textwidth}
    \centering
    \includegraphics[width=0.9\textwidth]{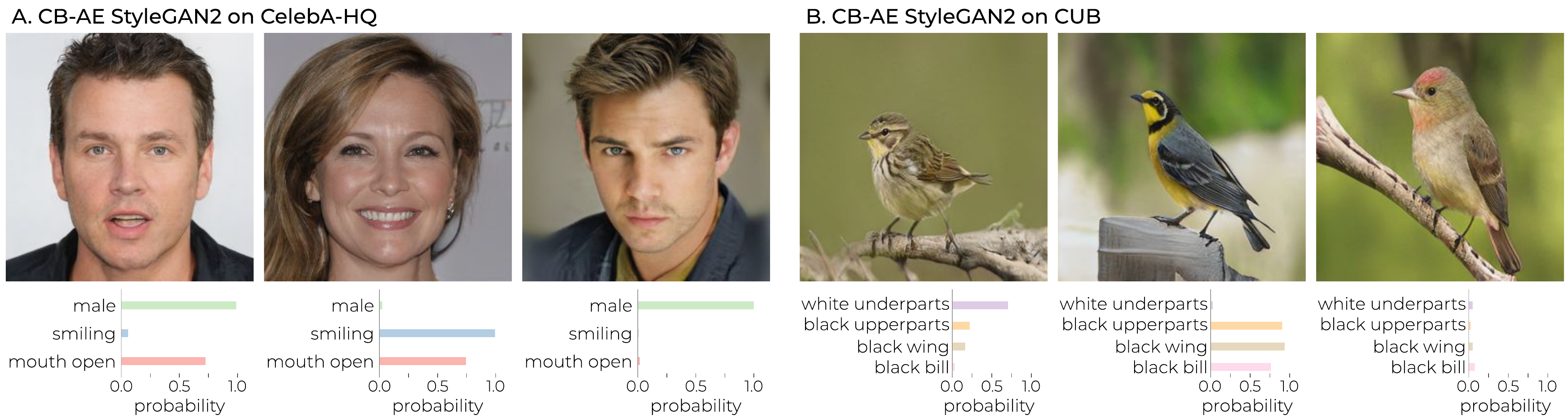}
    \vspace{-3mm}
    \caption{Samples generated using CB-AE with CelebA-HQ and CUB pretrained StyleGAN2 models along with concept probabilities.}
    \label{fig:cbae_conc_viz}
  \end{minipage}

  \vspace{1em}

  \begin{minipage}{\textwidth}
    \centering
    \includegraphics[width=\textwidth]{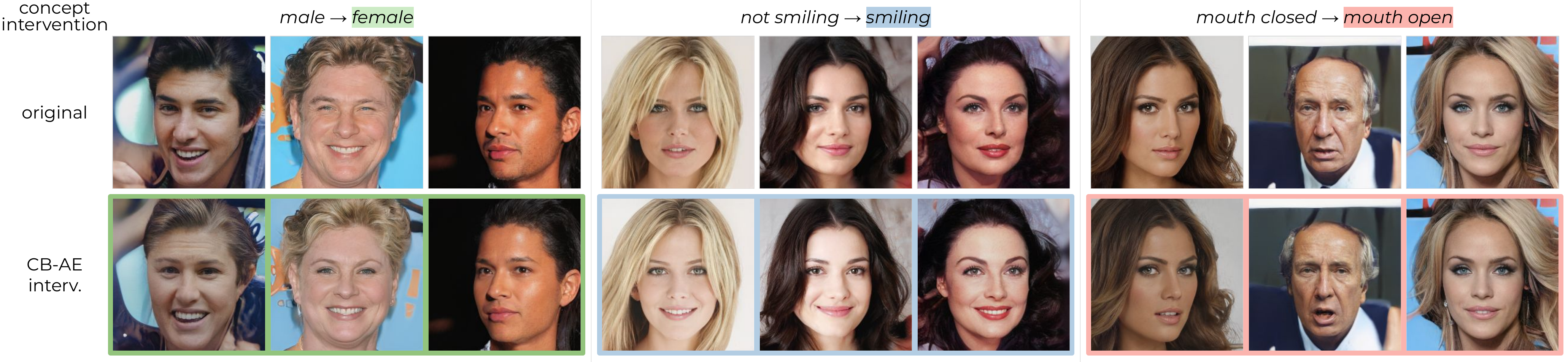}
    \vspace{-6mm}
    \caption{Concept intervention examples for CB-AE with CelebA-HQ pretrained StyleGAN2.}
    \label{fig:cbae_interv_viz}
  \end{minipage}

  \vspace{1em}

  \begin{minipage}{\textwidth}
    \centering
    \includegraphics[width=\textwidth]{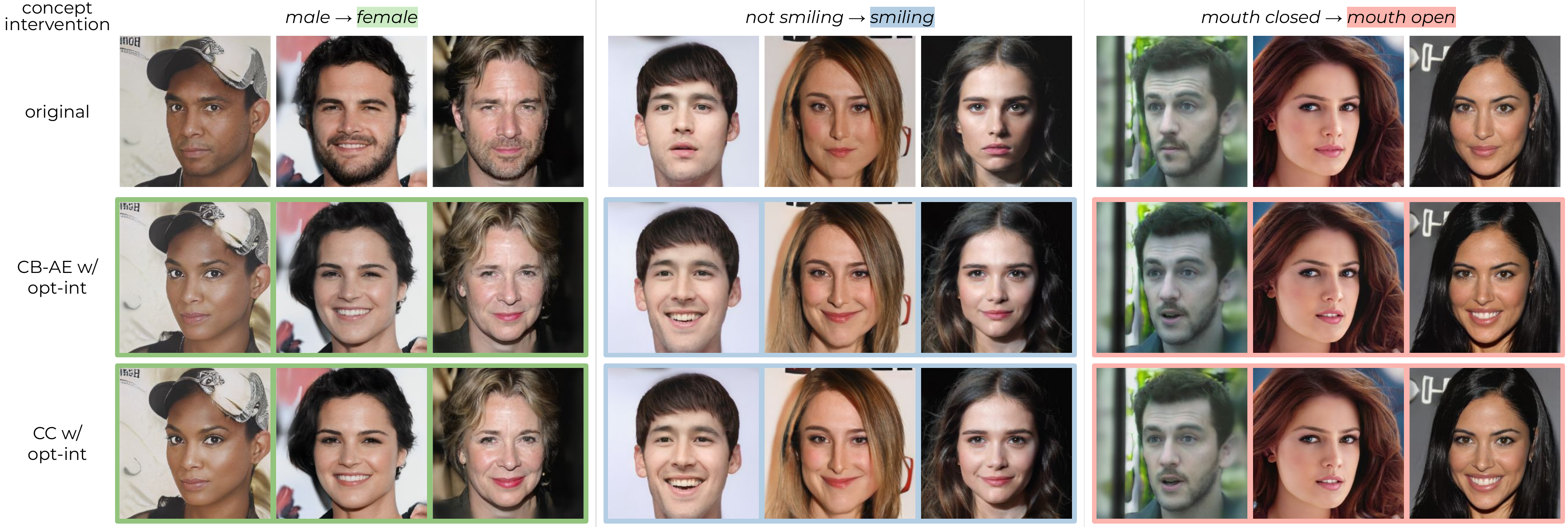}
    \vspace{-6mm}
    \caption{Optimization-based concept intervention (opt-int) examples for CB-AE and CC with CelebA-HQ pretrained StyleGAN2.}
    \label{fig:optint_viz}
  \end{minipage}
  \vspace{-4mm}

\end{figure*}

\subsection{Post-hoc Concept Controller (CC) for Steering}
\label{subsec:cc}

While optimization-based interventions work well empirically, it is interesting to note that the CB-AE decoder $D$ is not involved in the process. This leads us to question whether the CB-AE decoder can be removed if a user only plans to perform optimization-based interventions. While such a removal would not result in a CBM, it would be sufficient for the purpose of steering image generation. For this particular use case, it would be even more efficient than training the CB-AE since reconstruction and intervention losses are no longer required.
Hence, we propose a post-hoc concept controller (CC), denoted as $\Omega$, that predicts the concepts, \ie $c=\Omega(g_1(z))$ to efficiently steer image generation (Fig.\ \ref{fig:post_hoc_setup}\cvprcolor{B}). 

\noindent
\textbf{Training.}
For CC, the training is simply Objective 2 of the CB-AE training, \ie the cross-entropy loss $\mathcal{L}_c$ \wrt the concept pseudo-labels $\hat{y}=M(x)$ from the pseudo-label source $M$. Formally, the objective is $\min_{\Omega} [\mathcal{L}_c(\hat{y}, c)]$ where $c=\Omega(g_1(z))$ are the predicted concepts, and the loss encourages the concept controller $\Omega$ to align with the pseudo-label source $M$. Similar to the CB-AE training, we avoid the requirement of any real images for training as well as the requirement of concept labels.

\section{Experiments}

We detail the experimental setup and comprehensively evaluate our proposed methods with respect to the state-of-the-art prior work as well as other baselines.

\subsection{Experimental setup}

\noindent
\textbf{Base generative models and datasets.} 
We evaluate our CB-AE and CC methods on diverse generative models including GAN \cite{goodfellow2014generative}, Progressive GAN \cite{karras2018progressive}, StyleGAN2 \cite{karras2020analyzing}, and DDPM \cite{ho2020denoising}. We use models pretrained on standard datasets of varying image resolution ($64\times 64$ to $512\times 512$) like CelebA \cite{liu2015faceattributes}, CelebA-HQ \cite{lee2020maskgan}, and CUB \cite{WahCUB_200_2011}.
Following CBGM \cite{ismail2024concept}, we evaluate on the small balanced concept regime with the 8 most balanced concepts and the large unbalanced concept regime with all 40 concepts from the dataset. For CUB, we use 10 balanced concepts as per CBGM. Please refer to the Appendix for more details.

\noindent
\textbf{CB-AE and CC.}
We use a 4-layer MLP or 4 convolution (and transposed convolutional) layers for CB-AE encoder or CC (and decoder) depending on the dimensions of latent $w$. 
We use 
\revision{unsupervised concept embedding $\in \mathbb{R}^{40}$}
in CB-AE (ensuring bottleneck is much smaller than the latent).
As in CBGM \cite{ismail2024concept}, we train CB-AE/CC for 50 epochs with batch size 64.
For optimization-based interventions, we use 50-step I-RFGSM \cite{wong2020fast} with $\ell_\infty$-norm bound $\epsilon\!=\!0.1$. 
Please refer to the Appendix for complete implementation details.

\noindent
\textbf{Pseudo-label source $\boldsymbol{M}$.}
We consider three variants for $M$ with varying levels of concept supervision. First, we use off-the-shelf supervised (ResNet18-based) concept classifiers.
Second, with no concept supervision, we use CLIP zero-shot classifier \cite{radford2021learning} with only the concept names to obtain concept pseudo-labels. Third, as a compromise between the above two, we use TIP \cite{zhang2022tip} which is a few-shot-labeled version of CLIP zero-shot classifier, utilizing 128 concept-labeled real images. Unless otherwise mentioned, our experiments use the supervised classifiers for a fair comparison with the prior work CBGM \cite{ismail2024concept} that utilizes concept labels. 

\noindent
\textbf{Automated evaluation.} 
Following CBGM \cite{ismail2024concept}, we train concept classifiers (ViT-L-16-based) on real images and concept labels with high accuracy on a held-out test set. Note that these classifiers are separate and have higher accuracy than those used for pseudo-labels.
We evaluate our method using three automated metrics:
\begin{itemize}
    \itemsep0em
    \item \textbf{Concept Accuracy} is computed over 5k generated images as the average agreement between the supervised classifiers and our proposed CB-AE or CC.
    \item \textbf{Steerability} \cite{ismail2024concept}: 
    For each target concept, 
    we find 5k latents that do not produce the target concept (\ie probability of target concept $<0.5$ from the supervised concept classifier). For these latents, we perform the concept intervention using either the baselines or our methods, and compute the steerability as the percentage of intervened images that are classified to have the target concept.
    \item \textbf{Generation Quality} is evaluated using the standard Fréchet Inception Distance (FID) \cite{heusel2017gans}.
\end{itemize}
Intuitively, the concept accuracy, FID, and steerability metrics measure how well the concept, reconstruction, and intervention objectives, respectively, are satisfied.

\noindent
\textbf{Human evaluation.} We conduct a large-scale user study on Amazon Mechanical Turk to validate the automated evaluation of concept accuracy and steerability.
For both metrics, we display 10 images at a time and ask the user to click on images that match a displayed concept $c_i^+$ (see Appendix for more details). The images are collected as follows:
\begin{itemize}
    \itemsep0em
    \item \textbf{Concept Accuracy}: We collect and shuffle generated images and save their CB-AE and CC concept predictions ($c_i^+$ or $c_i^-$). 
    Based on user responses, we compute human agreement rate \wrt CB-AE/CC predictions.
    \item \textbf{Steerability}: We collect and shuffle generated images of concept $c_i^-$ with intervened images of concept $c_i^+$.
    Based on user responses, we compute human agreement rate on whether intervened images actually contain concept $c_i^+$. Shuffling the original and intervened images ensures that users are not biased towards clicking all images.
\end{itemize}
We evaluate using approximately 100 images per concept and per method for two concepts from CelebA-HQ: ``smiling" and ``male/female". Each set of 10 images is evaluated by 3 different users. Refer to the Appendix for more details.

\begin{table*}[t]
\caption{\textbf{Concept accuracy evaluation} on 5k samples with 8 concepts for CelebA, CelebA-HQ, and 10 concepts for CUB. }
\label{tab:conc_acc}
\vspace{-3mm}
\setlength{\tabcolsep}{3.5mm}
\resizebox{\linewidth}{!}{
\begin{tabular}{@{}lcccccccccc@{}}
\toprule
\multicolumn{1}{c}{\multirow{2}{*}{Conc.\ Acc.\ (\%)}} &
  \multicolumn{3}{c}{CelebA (64$\times$64)} &
  \multicolumn{3}{c}{CelebA-HQ (256$\times$256)} &
  \multicolumn{2}{c}{CUB (64$\times$64)} &
  CUB (256$\times$256) \\ \cmidrule(l){2-4} \cmidrule(l){5-7} \cmidrule(l){8-9} \cmidrule(l){10-10} 
\multicolumn{1}{c}{} &
  GAN &
  PGAN &
  DDPM &
  DDPM &
  StyGAN2 &
  PGAN &
  DDPM &
  GAN &
  StyGAN2 \\ \midrule
  \textit{Ours} (CB-AE) & 86.56 & 87.87 & 84.98 & 89.79 & \textbf{86.04} & 82.68 & 72.35 & 74.41 & \textbf{81.33} \\
\textit{Ours} (CC)  & \textbf{87.65} & \textbf{90.00} & \textbf{85.13} & \textbf{89.82} & 83.57 & \textbf{83.94} & \textbf{72.87} & \textbf{75.60} & 81.11  \\ 
\bottomrule
\end{tabular}
}
\end{table*}

\begin{table*}[t]
\caption{\textbf{Extended steerability evaluation} on 5k samples with 8 concepts for CelebA, CelebA-HQ, and 10 concepts for CUB. $^\dagger$CBGM numbers are from their paper (1k samples) 
since their results are not reproducible using their released code. For CBGM training time, we used the time taken by their code for GAN and base model training times for other models (whose code was not available) with 1 V100 GPU.
}
\label{tab:int_succ_rate}
\vspace{-3mm}
\setlength{\tabcolsep}{3.5mm}
\resizebox{\linewidth}{!}{
\begin{tabular}{@{}llcccccccccc@{}}
\toprule
\multicolumn{1}{c}{\multirow{2}{*}{Steerability (\%)}} & &
  \multicolumn{3}{c}{CelebA (64$\times$64)} &
  \multicolumn{3}{c}{CelebA-HQ (256$\times$256)} &
  \multicolumn{2}{c}{CUB (64$\times$64)} &
  CUB (256$\times$256) \\ \cmidrule(l){3-5} \cmidrule(l){6-8} \cmidrule(l){9-10} \cmidrule(l){11-11} 
\multicolumn{1}{c}{} & &
  GAN &
  PGAN &
  DDPM &
  DDPM &
  StyGAN2 &
  PGAN &
  DDPM &
  GAN &
  StyGAN2 \\ \midrule
CBGM \cite{ismail2024concept}$^\dagger$ & &
  25.60 &
  - &
  13.80 &
  - &
  - &
  - &
  14.80 &
  21.30 &
  - \\
  \textit{Ours} (CB-AE) & &
   {47.34} &
   {40.31} &
   {23.72} &
   {23.50} &
   40.27 &
   {29.31} &
   {25.64} & 
   {20.89} &
   {10.52}
  \\
  \textit{Ours} (CB-AE+opt-int) & &
  \textbf{{61.14}} &
  {41.73} &
  {38.09} &
  {50.49} &
  61.66 &
  {32.10} &
  {36.94} &
  {46.03} &
  \textbf{{65.11}}
  \\
\textit{Ours} (CC+opt-int) & &
  51.14 &
  \textbf{58.94} &
  \textbf{{41.45}} &
  \textbf{{56.70}} &
  \textbf{{67.95}} &
  \textbf{47.29} &
  \textbf{{44.47}} &
  \textbf{{48.91}} &
  44.72 \\ 

\midrule
\multirow{2}{*}{\makecell{Train time reduction \\ \wrt CBGM}}{} & CB-AE & 3.7$\times$ & 5.4$\times$ & 4$\times$ & 8.1$\times$ & 3.5$\times$ & 2$\times$ & 3.7$\times$ & 3.3$\times$ & 3.1$\times$ \\
 & CC & 30.9$\times$ & 20.3$\times$ & 10$\times$ & 28.9$\times$ & 8.3$\times$ & 6.7$\times$ & 8.5$\times$ & 21$\times$ & 7.1$\times$ \\
\bottomrule
\end{tabular}
}
\end{table*}

\vspace{-1mm}
\subsection{Evaluation}
\vspace{-1mm}

\noindent
\textbf{Qualitative evaluation.}
In Fig.\ \ref{fig:cbae_conc_viz}, we present CB-AE-StyleGAN2 generated images for CelebA-HQ and CUB datasets with corresponding concept predictions. We also visualize CB-AE interventions in Fig.\ \ref{fig:cbae_interv_viz} and optimization-based interventions in Fig.\ \ref{fig:optint_viz}. Interestingly, with the same latent $w$, both CB-AE and CC lead to very similar optimization-based interventions, which is reasonable since the same pseudo-label source $M$ was used in both cases. Overall, we find that the optimization-based interventions are relatively higher quality and more orthogonal (\ie less changes to other concepts) than the CB-AE interventions.

\noindent
\textbf{Concept Accuracy.} 
In Table \ref{tab:conc_acc}, we report the concept accuracies for our CB-AE and CC. We could not compare with CBGM since they do not evaluate concept accuracy, and since we could not reproduce their results. Overall, we find that both CB-AE and CC achieve good concept accuracies across various datasets, models and image resolutions. 

In most scenarios, CC outperforms CB-AE since concept alignment is the sole objective optimized in CC, while CB-AE has additional objectives.
However, we observe that CB-AE outperforms CC only for StyleGAN2. This is because the multiple objectives in CB-AE training may be easier to balance when dealing with clean latents (GANs) than with noisy latents (DDPM), leading to CB-AE outperforming CC for StyleGAN2. This is supported by average loss for CB-AE being lower for StyleGAN2 (0.68) than DDPM (0.92).

\begin{table}[]
\centering
\caption{\textbf{Steerability comparisons} with CBGM \cite{ismail2024concept} and other baseline intervention methods computed on 1k samples (each experiment repeated three times for mean and standard deviation).
}
\vspace{-3mm}
\label{tab:cbgm_comparisons}
\setlength{\tabcolsep}{0.5mm}
\resizebox{\linewidth}{!}{
\begin{tabular}{lccc}
\toprule
\textbf{Concept Regime}        & \multicolumn{2}{c}{Small balanced concepts} & Large unbalanced concepts \\ 
\cmidrule(lr){2-3}\cmidrule(lr){4-4}
\textbf{Dataset}    & CUB (10 conc.) & CelebA (8 conc.) & CelebA (40 conc.) \\ \midrule
\multicolumn{4}{l}{\textbf{Baseline Intervention Methods}} \\
CGAN \cite{mirza2014conditional}      & 5.4 $\pm$ 0.4  & 8.7 $\pm$ 1.3    & 2.9 $\pm$ 0.0     \\
ACGAN \cite{odena2017conditional}     & 18.5 $\pm$ 0.4 & 9.2 $\pm$ 0.7    & 1.2 $\pm$ 0.1     \\
CB-GAN \cite{ismail2024concept}     & 21.3 $\pm$ 0.3 & 25.6 $\pm$ 0.5   & 23.1 $\pm$ 0.2    \\
\midrule
\textbf{Our Methods} \\
CB-AE-GAN             & 19.9 $\pm$ 0.9 & 47.1 $\pm$ 0.5 &  45.5 $\pm$ 0.4     \\
CB-AE-GAN+opt-int  & 45.1 $\pm$ 1.1 & \textbf{59.9 $\pm$ 1.8} & \textbf{58.3 $\pm$ 1.3}      \\
CC-GAN+opt-int     & \textbf{49.3 $\pm$ 0.6} & 50.8 $\pm$ 1.2 & 49.7 $\pm$ 0.9      \\ \midrule \midrule
\multicolumn{4}{l}{\textbf{Baseline Intervention Methods}} \\
CF-DDPM \cite{ho2021classifier}   & 2.7 $\pm$ 1.9  & 7.2 $\pm$ 3.8    & 5.1 $\pm$ 2.4     \\
CG-DDPM \cite{dhariwal2021diffusion}   & 2.1 $\pm$ 1.4  & 6.8 $\pm$ 1.1    & 5.4 $\pm$ 2.6     \\
CB-DDPM \cite{ismail2024concept}   & 14.8 $\pm$ 6.2 & 13.8 $\pm$ 2.7   & 12.6 $\pm$ 1.7    \\
\midrule
\textbf{Our Methods} \\
CB-AE-DDPM &   25.8 $\pm$ 1.1             &  23.1 $\pm$ 0.9      & 23.6 $\pm$ 1.0                  \\
CB-AE-DDPM+opt-int & 37.3 $\pm$ 1.5 & 37.5 $\pm$ 1.3 &   36.2 $\pm$ 1.1    \\
CC-DDPM+opt-int    & \textbf{45.4 $\pm$ 2.2} & \textbf{41.8 $\pm$ 1.8} &  \textbf{41.3 $\pm$ 1.5}    \\ 
\bottomrule
\end{tabular}
}
\end{table}

\noindent
\textbf{Steerability.}
We compare the concept steerability of our methods on 1k samples
with GAN intervention methods like conditional GAN (CGAN) \cite{mirza2014conditional}, auxiliary classifier GAN (ACGAN) \cite{odena2017conditional}, and CBGM (CB-GAN) \cite{ismail2024concept}, and with diffusion model intervention methods like classifier-guided (CG) \cite{dhariwal2021diffusion} DDPM, classifier-free (CF) \cite{ho2021classifier} DDPM, and concept bottleneck DDPM (CB-DDPM) \cite{ismail2024concept}
in Table \ref{tab:cbgm_comparisons}. We observe significant gains over CBGM \cite{ismail2024concept} for both GAN (average {+14.2}\% for CB-AE, {+31.1}\% for CB-AE with opt-int, and {+26.6}\% for CC) and for DDPM (average {+10.4}\% for CB-AE, {+23.3}\% for CB-AE with opt-int, and {+29.1}\% for CC). 

Table \ref{tab:int_succ_rate} presents extended steerability evaluation on 5k samples with other types of GANs like Progressive GAN \cite{karras2018progressive} and StyleGAN2 \cite{karras2020analyzing} for higher resolution (256$\times$256) datasets. 
We could not obtain CBGM's results for these settings using their code.
Overall, we find optimization-based interventions (opt-int) outperform the CB-AE intervention method, and CC generally outperforms CB-AE. 
\revision{Intuitively, opt-int involves instance-specific and iterative optimization, while CB-AE is applied in the same way to all samples.}
Although CB-AE is trained with intervention losses, its steerability tends to be worse than CC since CB-AE is more challenging to train, with multiple objectives to satisfy.

\begin{table}[]
\centering
\caption{\textbf{Generation quality and training time comparisons} for CelebA-HQ with StyleGAN2. $^\dagger$CBGM results are from their paper.
Training time is in V100 GPU-hours.
}
\label{tab:gen_quality}
\vspace{-3mm}
\resizebox{\linewidth}{!}{
\begin{tabular}{lccc}
\toprule
\multicolumn{1}{c}{FID ($\downarrow$)}   & \multicolumn{1}{c}{CBGM$^\dagger$ \cite{ismail2024concept}} & \multicolumn{1}{c}{CB-AE (\textit{Ours})} & CC (\textit{Ours}) \\
\midrule
Base model                             & 9.0           & 7.66   & 7.66  \\
CB model                           & 9.1           & 9.52  & -   \\
CB interv.                             & -             & 9.65  & -    \\
Opt-interv.                           & -             & 7.67 & 7.65   \\
\midrule
Train time (hrs) & 50 & 14 & 6 \\
\bottomrule
\end{tabular}
}
\end{table}

\noindent
\textbf{Generation quality.} 
We compare the generation quality of CBGM \cite{ismail2024concept} with CB-AE and CC in Table \ref{tab:gen_quality}. For CB-AE, we observe a relatively higher drop in image quality than CBGM, but our methods are trained 3.5-8$\times$ faster, and do not require training from scratch. For a high-quality StyleGAN2 model, our CB-AE and CC with optimization-based interventions can produce almost the same quality of images as the base model while having high steerability (61.66\% and 67.95\% respectively, from Table \ref{tab:int_succ_rate}). 

Intuitively, since CB-AE focuses on concept and intervention losses, we obtain better steerability, while CBGM has better FID since they include generative model losses. Hence, there is a tradeoff between image quality and interpretability which can be improved in future work.

\noindent
\textbf{Human evaluation.} 
In Table \ref{tab:human_eval}, we compare human agreement rate with automated evaluation of concept accuracy and steerability. We chose easily recognizable concepts, ``gender" and ``smiling" as representative of low and high steerability respectively. Overall, our automated evaluation is similar to human agreement (with room for improving the classifiers), validating the usefulness of automated evaluation.

\begin{table}[]
\centering
\caption{\textbf{Human evaluation results} for CB-AE and CC trained with CelebA-HQ pretrained StyleGAN2.}
\vspace{-3mm}
\label{tab:human_eval}
\begin{minipage}{\linewidth}
\resizebox{\linewidth}{!}{
\begin{tabular}{lcccc}
\toprule
\multirow{2}{*}{Conc.\ Acc.\ (\%)} & \multicolumn{2}{c}{Smiling} & \multicolumn{2}{c}{Male/Female} \\ \cmidrule(lr){2-3}\cmidrule(lr){4-5}
 & Automated & Human & Automated & Human \\
 \midrule
CB-AE (\textit{Ours}) & 92.38     & 86.35 & 100.0     & 94.06 \\
CC (\textit{Ours})    & 89.47     & 80.35 & 96.96     & 96.30 \\
\bottomrule
\end{tabular}
}
\end{minipage}

\vspace{1mm}

\begin{minipage}{\linewidth}
\setlength{\tabcolsep}{1mm}
\resizebox{\linewidth}{!}{
\begin{tabular}{lcccc}
\toprule
\multirow{2}{*}{Steerability (\%)} & \multicolumn{2}{c}{Smiling} & \multicolumn{2}{c}{Female} \\ \cmidrule(lr){2-3}\cmidrule(lr){4-5}
& Automated & Human & Automated & Human \\ \midrule
CB-AE (\textit{Ours})            & 65.90     & 77.27 & 17.02     & 17.73 \\
CB-AE w/ opt-int (\textit{Ours}) & 76.59     & 78.72 & 42.85     & 41.50 \\
CC w/ opt-int (\textit{Ours})    & 77.36     & 77.36 & 26.92     & 19.87 \\
\bottomrule
\end{tabular}
}
\end{minipage}
\end{table}

\subsection{Analysis}
\vspace{-1mm}

\noindent
\textbf{Ablation study.}
We analyze the contribution of each
CB-AE training loss.
However, we do not ablate concept loss $\mathcal{L}_c$ or latent reconstruction loss $\mathcal{L}_{r_1}$ since our evaluation metrics would be meaningless
if the CB-AE cannot predict the concepts or cannot reconstruct the generator latent $w'$. 
Hence, Table \ref{tab:ablation} ablates image reconstruction loss $\mathcal{L}_{r_2}$ from Eq.\ \eqref{eqn:recon_loss}, intervened concept loss $\mathcal{L}_{i_1}$ from Eq.\ \eqref{eqn:int_loss1}, and cyclic intervened concept loss $\mathcal{L}_{i_2}$ from Eq.\ \eqref{eqn:int_loss2}. The ablations are for the most challenging pseudo-label setting of $M$ (CLIP-zero-shot), since it is most affected by loss ablations.

In Table \ref{tab:ablation}, we observe that using the image reconstruction loss $\mathcal{L}_{r_2}$ improves the generation quality FID (row \#2 vs.\ \#1), which is intuitive since this loss directly encourages the images to be closer to the original ones.
Next, using only the intervened concept loss or only the intervened cyclic loss improves the steerability while trading-off generation quality (row \#3 vs.\ \#2 or row \#4 vs.\ \#2). 
Finally, using both intervened losses significantly improves both concept accuracy and steerability (row \#5 vs. \#3 and \#4), while generation quality remains similar. Overall, both intervention losses are crucial to ensure good concept accuracy and steerability, while image reconstruction loss improves generation quality. 

\noindent
\textbf{Sensitivity to pseudo-label source $\boldsymbol{M}$.} 
Table \ref{tab:pl_sens} compares concept accuracy and steerability when varying pseudo-label source $M$. Concept accuracy improves as pseudo-label quality improves from CLIP-zero-shot to supervised classifiers. While TIP few-shot does not improve CB-AE interventions, it significantly improves optimization-based interventions and highlights the usefulness of even limited labeled data.
\revision{Further, our method can use newer CLIP models like SigLIP \cite{zhai2023sigmoid}, OpenCLIP \cite{cherti2023reproducible} to further improve performance.}

\begin{table}[]
\centering
\caption{\textbf{Ablation study on CB-AE training objectives} for the most challenging CLIP-zero-shot pseudo-label setting for CelebA-HQ pretrained StyleGAN2. 
$\mathcal{L}_{r_2}, \mathcal{L}_{i_1}, \mathcal{L}_{i_2}$ indicate image reconstruction loss, intervened concept loss, and intervened cyclic loss respectively from Eq.\ \eqref{eqn:recon_loss}, \eqref{eqn:int_loss1}, \eqref{eqn:int_loss2}.}
\vspace{-3mm}
\label{tab:ablation}
\resizebox{\linewidth}{!}{
\begin{tabular}{ccccccc}
\toprule
\multirow{2}{*}{\begin{tabular}[c]{@{}c@{}}\\[-2ex]Row \\ \# \end{tabular}}& \multirow{2}{*}{$\mathcal{L}_{r_2}$} & \multirow{2}{*}{$\mathcal{L}_{i_1}$} & \multirow{2}{*}{$\mathcal{L}_{i_2}$} & \multicolumn{3}{c}{Trained with $M\!=\!$ CLIP-zero-shot} \\
\cmidrule{5-7}
 & &  &  & Conc.\ Acc.\ (\%) & Steerability (\%) & FID ($\downarrow$) \\
 \midrule
1 & \xmark & \xmark & \xmark &  58.87               &    6.01               &  11.06                  \\
2 & \cmark & \xmark & \xmark & 59.32                &  6.51                 &  \textbf{9.88}                  \\
3 & \cmark & \cmark & \xmark & 57.75                &  9.98                 & 12.42                   \\
4 & \cmark & \xmark & \cmark & 58.33                &   13.03                &  13.62                  \\
5 & \cmark & \cmark & \cmark & \textbf{67.26}                & \textbf{20.61}                  & 12.82                   \\
\bottomrule
\end{tabular}
}
\end{table}

\begin{table}[]
\caption{\textbf{Sensitivity to pseudo-label source $\boldsymbol{M}$}. For CB-AE trained with CelebA-HQ pretrained StyleGAN2, we compare concept accuracy and steerability with different $M$: CLIP-zero-shot \cite{radford2021learning}, TIP-few-shot \cite{zhang2022tip}, or supervised classifiers.}
\label{tab:pl_sens}
\setlength{\tabcolsep}{3.5mm}
\vspace{-3mm}
\resizebox{\linewidth}{!}{
\begin{tabular}{lcccc}
\toprule
\multirow{2}{*}{\begin{tabular}[c]{@{}l@{}}Pseudo-label \\ source $M$\end{tabular}} &
  \multirow{2}{*}{\begin{tabular}[c]{@{}c@{}}Conc. Acc. \\ (\%)\end{tabular}} &
  \multicolumn{2}{c}{Steerability (\%)} \\ \cmidrule(l){3-4} 
            &       & CB-AE & CB-AE w/ opt-int \\ \midrule
    CLIP-zs & 67.26 & 20.61 & 29.23 \\
    TIP-fs-128  & 76.08 & 21.51 & 38.73 \\
    Supervised-clsf & \textbf{86.04} & \textbf{40.27} & \textbf{61.66} \\    
\bottomrule
\end{tabular}
}
\end{table}

\vspace{-2mm}
\section{Conclusion}
\vspace{-1mm}

In this work, we proposed two novel and low-cost methods, concept-bottleneck autoencoder (CB-AE) and concept controller (CC), to efficiently build interpretable generative models from pretrained models. Compared to the prior approach that struggles with efficiency and scalability, our methods achieve 4-15$\times$ faster training, require minimal to no concept supervision, and generalize across modern generative model families including GANs and diffusion models with 25\% improved steerability on average.

\section*{Acknowledgements}
This work is supported in part by National Science Foundation (NSF) awards CNS-1730158, ACI-1540112, ACI-1541349, OAC-1826967, OAC-2112167, CNS-2100237, CNS-2120019, the University of California Office of the President, and the University of California San Diego's California Institute for Telecommunications and Information Technology/Qualcomm Institute. This work used Delta CPU, GPU and Storage through allocation CIS230153 from the Advanced Cyberinfrastructure Coordination Ecosystem: Services \& Support program, which is supported by National Science Foundation grants 2138259, 2138286, 2138307, 2137603, and 2138296. The authors are partially supported by National Science Foundation under Grant No. 2107189, 2313105, 2430539, Hellman Fellowship, and Intel Rising Star Faculty Award. The authors would also like to thank anonymous reviewers for valuable feedback to improve the manuscript.

\appendix

\section*{Appendix}

\noindent
In this appendix, we provide comprehensive implementation details and more analysis experiments. Towards reproducible research, we will release our complete codebase and pretrained weights.
The appendix is organized as follows:

\renewcommand{\labelitemii}{$\circ$}

\begin{itemize}
\setlength{\itemindent}{-0mm}
    \item Section~\ref{sup:sec:limitations}: Limitations
    \item Section~\ref{sup:sec:implementation}: Implementation Details
    \begin{itemize}
        \setlength{\itemindent}{-0mm}
        \item Datasets (Sec.\ \ref{sup:subsec:datasets})
        \item Architecture details (Sec.\ \ref{sup:subsec:arch})
        \item Training details (Sec.\ \ref{sup:subsec:training_details})
        \item Human evaluation details (Sec.\ \ref{sup:subsec:humaneval_details}, Fig.\ \ref{sup:fig:mturk_example})
        \item Miscellaneous details (Sec.\ \ref{sup:subsec:compute_details})
    \end{itemize}
    \item Section~\ref{sup:sec:expts}: Experiments
    \begin{itemize}
        \setlength{\itemindent}{-0mm}
        \item Extended comparisons (Sec.\ \ref{sup:subsec:extended_comp}, Table \ref{sup:tab:perconc_steerability})
        \item Extended analysis (Sec.\ \ref{sup:subsec:extended_analysis}, Table \ref{sup:tab:ablation}-\ref{tab:scaling_num_conc}, Fig.\ \ref{sup:fig:optint_sens}-\ref{sup:fig:conc_interp_examples})
        \item Efficiency analysis (Sec.\ \ref{sup:subsec:efficiency_analysis}, Table \ref{sup:tab:efficiency}, Table \ref{sup:tab:trainable_params})
    \end{itemize}
\end{itemize}{}

\section{Limitations}
\label{sup:sec:limitations}

While the steerability metric quantifies whether the target concept is obtained in the intervened image, it does not quantify if other concepts (outside the known concepts) have changed. For example, an intervention from ``not smiling" to ``smiling" may lead to a smiling image with different hair color. This cannot be easily identified with an automated metric, and it is challenging and expensive to design an unbiased human evaluation given its subjective nature. It will be interesting to address this in future work.

\section{Implementation Details}
\label{sup:sec:implementation}

\subsection{Datasets}
\label{sup:subsec:datasets}

For the CelebA dataset, we follow CBGM \cite{ismail2024concept} and use 8 balanced concepts for the balanced concept regime. We determine these concepts based on the fraction of number of images that contain a particular concept \wrt number of images that do not contain that concept. The 8 concepts for CelebA are ``smiling", ``male", ``heavy makeup", ``mouth open", ``attractive", ``wearing lipstick", ``high cheekbones", and ``wavy hair".
For CelebA-HQ, we have the same 8 concepts with the exception of ``wavy hair", which is replaced by ``arched eyebrows".
For CUB dataset, we use the 10 most balanced concepts: ``small size (5 to 9 inches)", ``perching-like shape", ``solid breast pattern", ``black bill color", ``bill length shorter than head", ``black wing color", ``solid belly pattern", ``all purpose bill shape", ``black upperparts color", and ``white underparts color", following CBGM \cite{ismail2024concept}.
For the steerability metric, we consider 16 and 20 target concepts for CelebA and CUB respectively since they are binary concepts.

\subsection{Architecture details}
\label{sup:subsec:arch}

For base generative models with vector latents or small spatial latents like StyleGAN2 or DDPM, we use a 4-layer MLP (with batch norm and leaky ReLU) for both CB-AE encoder $E$ and decoder $D$. For models with larger spatial latents like GAN or PGAN, we use 4 convolution (and transposed convolution) layers with batch norm and leaky ReLU for the CB-AE encoder $E$ (and decoder $D$). CC has the same architecture as the CB-AE encoder $E$.

\subsection{Training details}
\label{sup:subsec:training_details}

For GANs, we use the training procedure as detailed in the main paper. For the DDPM diffusion model, we use saved generated images instead of generating the images at training time since DDPM generation is relatively slower than GANs. Further, we follow the diffusion model noising procedure where,
at each training iteration, we choose a random timestep $t$ and add the corresponding level of noise to the generated image before passing it through first part of the generative model $g_1$ (UNet encoder for DDPM). Since the CB-AE/CC would be used at different steps of denoising, it is trained using noised latents (instead of only clean latents from clean images). For GANs, $g_2$ produces an image while DDPM's $g_2$ predicts the estimated noise. So, we use the initial clean image to obtain the pseudo-label from $M$ instead of the output of $g_2$. Apart from this, we follow the same training procedure as discussed in the main paper. While we use the noising techniques from DDPM, the training losses of DDPM are not used and only the CB-AE/CC is trained with our proposed losses.

While in original DDPM training, $t$ is chosen from 0 (clean image) to 999 (complete noise), we restrict the choice of $t$ from 0 to 400, similar to \cite{gandikota2024sliders}. This is because the CB-AE has to predict the concepts and in practice, the generated images are very noisy at $t>400$. 

Based on this, at inference time, we use the CB-AE only for $t<400$ and use the base model for $t>400$. We also use the 50-step DDIM sampler \cite{song2021denoising} at inference time instead of the DDPM sampler since it is much faster with similar image quality. Note that DDIM converts the 1000 steps into 50 steps but retains the range of $t$ from 0 to 999.

\begin{figure}[t]
    \centering
    \includegraphics[width=\linewidth]{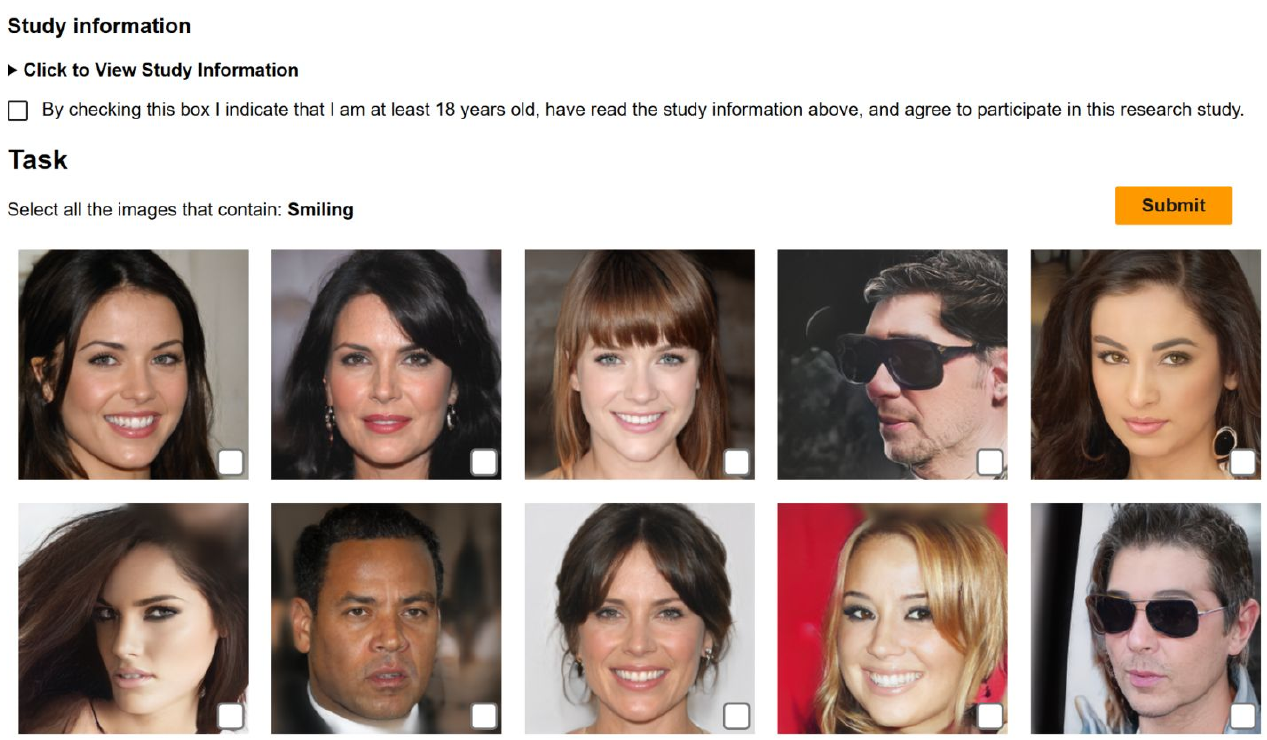}
    \caption{User interface shown to Amazon Mechanical Turk users. We ask users to click on images which match the displayed concept.}
    \label{sup:fig:mturk_example}
\end{figure}

\begin{table}[]
\centering
\caption{\textbf{Per-concept steerability comparison} on CelebA dataset. Results for baseline intervention methods are from CBGM \cite{ismail2024concept}. Note that average results in the main paper are over 16 target concepts, but here we compare with the available CBGM results.}
\vspace{-3mm}
\label{sup:tab:perconc_steerability}
\setlength{\tabcolsep}{1mm}
\resizebox{\linewidth}{!}{
\begin{tabular}{lccccc}
\toprule
Concept & High Cheekbones & Male & Mouth Open & Smiling & Wavy Hair \\
\midrule
\multicolumn{6}{l}{\textbf{Baseline Intervention Methods}} \\
CGAN \cite{mirza2014conditional} & 5.8 & 6.0 & 6.1 & 3.6 & 13.5 \\
ACGAN \cite{odena2017conditional} & 11.8 & 9.3 & 13.5 & 14.3 & 8.4 \\
CB-GAN \cite{ismail2024concept} & 9.8 & 53.7 & 8.2 & 25.8 & 30.5 \\
\midrule
\textbf{Our Methods} \\
CB-AE-GAN & 48.1 & 35.0 & 51.3 & 64.5 & 27.6 \\
CB-AE-GAN+opt-int & \textbf{66.0} & \textbf{72.3} & \textbf{81.3} & \textbf{67.3} & \textbf{38.1} \\
CC-GAN+opt-int & 50.9 & 54.8 & 78.5 & 53.8 & 23.4 \\
\midrule
\midrule
\multicolumn{6}{l}{\textbf{Baseline Intervention Methods}} \\
CF-DDPM \cite{ho2021classifier} & 8.3 & 10.2 & 7.2 & 7.1 & 3.8 \\
CB-DDPM \cite{ismail2024concept} & 11.7 & 14.8 & 13.9 & 15.1 & 10.3 \\
\midrule
\textbf{Our Methods} \\
CB-AE-DDPM & 15.7 & 39.6 & 34.9 & 29.8 & 21.0 \\
CB-AE-DDPM+opt-int & 51.3 & \textbf{51.4} & \textbf{73.5} & 58.8 & 45.4 \\
CC-DDPM+opt-int & \textbf{61.9} & 42.6 & 63.3 & \textbf{64.0} & \textbf{65.9} \\
\bottomrule
\end{tabular}
}
\end{table}

\subsection{Human evaluation details}
\label{sup:subsec:humaneval_details}

For our user study on Amazon Mechanical Turk to validate the automated evaluation of concept accuracy and steerability, we display 10 images at a time and ask the user to click on images that match a displayed concept $c_i^+$, as shown in Fig.\ \ref{sup:fig:mturk_example}. To ensure the quality of user responses, we require users to be in the United States, have $>$ 98\% approval rate, and $>$ 10000 previously approved responses. For each set of 10 images, a user is paid \$0.05.

\subsection{Miscellaneous details}
\label{sup:subsec:compute_details}

We implement our framework in PyTorch \cite{paszke2019pytorch}.
For all experiments, we use 10 CPU cores, 90 GB RAM, and a single Nvidia Tesla V100 GPU with 32 GB VRAM.

\section{Experiments}
\label{sup:sec:expts}

\subsection{Extended comparisons}
\label{sup:subsec:extended_comp}

We present extended per-concept steerability comparisons with CBGM \cite{ismail2024concept} and other baseline intervention methods in Table \ref{sup:tab:perconc_steerability}. We compare the steerability on CelebA for the 5 concepts (out of 8) which are provided in the CBGM paper and find consistent improvements across all concepts.

\begin{table}[]
\centering
\caption{\textbf{Ablation study on CB-AE training objectives} for the supervised classifier pseudo-label setting for CelebA-HQ pretrained StyleGAN2. Concept loss $\mathcal{L}_c$ and latent reconstruction loss $\mathcal{L}_{r_1}$ are not ablated since they are essential to concept prediction and AE reconstruction. $\mathcal{L}_{r_2}, \mathcal{L}_{i_1}, \mathcal{L}_{i_2}$ indicate image reconstruction loss, intervened concept loss, and intervened cyclic loss respectively from Eq.\ \cvprcolor{1}, \cvprcolor{3} (main paper).}
\vspace{-3mm}
\label{sup:tab:ablation}
\resizebox{\linewidth}{!}{
\begin{tabular}{ccccccc}
\toprule
\multirow{2}{*}{\begin{tabular}[c]{@{}c@{}}\\[-2ex]Row \\ \# \end{tabular}}& \multirow{2}{*}{$\mathcal{L}_{r_2}$} & \multirow{2}{*}{$\mathcal{L}_{i_1}$} & \multirow{2}{*}{$\mathcal{L}_{i_2}$} & \multicolumn{3}{c}{Trained with $M\!=\!$ Supervised classifiers} \\
\cmidrule{5-7}
 & &  &  & Conc.\ Acc.\ (\%) & Steerability (\%) & FID ($\downarrow$) \\
 \midrule
1 & \xmark & \xmark & \xmark & 85.36 & 33.41 & 15.18 \\
2 & \cmark & \xmark & \xmark & 83.40 & 38.68 & 11.27 \\
3 & \cmark & \cmark & \xmark & 83.16 & 38.84 & 12.72 \\
4 & \cmark & \xmark & \cmark & 86.52 & 36.95 & 18.57 \\
5 & \cmark & \cmark & \cmark & 86.04 & 40.27 & 9.52 \\
\bottomrule
\end{tabular}
}
\end{table}
\begin{table}[]
\centering
\caption{\textbf{Steerability comparison when scaling image resolution} for our methods with PGAN and CelebA-HQ dataset.}
\vspace{-3mm}
\label{sup:tab:pgan_scaling_resolution}
\resizebox{\linewidth}{!}{
\begin{tabular}{cccc}
\toprule
Image Resolution & CB-AE & CB-AE+opt-int & CC+opt-int \\
\midrule
256$\times$256 & 29.31 & 32.10 & 47.29 \\
512$\times$512 & 26.48 & 34.92 & 36.87 \\
\bottomrule
\end{tabular}
}
\end{table}

\begin{figure*}
    \centering
    \includegraphics[width=\linewidth]{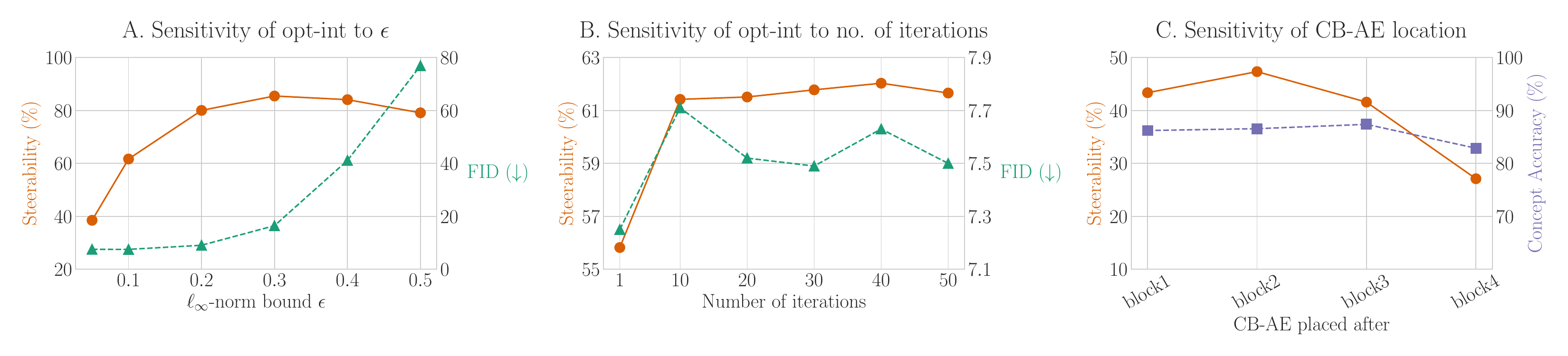}
    \vspace{-3mm}
    \caption{\textbf{A, B.} Sensitivity analysis of optimization-based interventions with CB-AE for CelebA-HQ, StyleGAN2 \wrt $\ell_\infty$-norm bound $\epsilon$ and number of iterations used in the optimization. 
    \revision{\textbf{C.} Sensitivity analysis of CB-AE location in GAN.}
    Note that orange circles represent steerability, green triangles represent FID, \revision{and purple squares represent concept accuracy}.}
    \label{sup:fig:optint_sens}
\end{figure*}

\subsection{Extended analysis}
\label{sup:subsec:extended_analysis}

\noindent
\textbf{Ablation study.} In the main paper, we performed the ablation study on CB-AE training objectives for the more challenging CLIP-zero-shot pseudo-label setting. In Table \ref{sup:tab:ablation}, we perform the same ablation study when using supervised classifiers as the pseudo-label source $M$. Similar to the results in the main paper, using the image reconstruction loss $\mathcal{L}_{r_2}$ leads to lower concept accuracy, higher steerability and better image quality (row \#2 vs.\ \#1, Table \ref{sup:tab:ablation}). Additionally using the intervened concept loss $\mathcal{L}_{i_1}$ improves the steerability and image quality but reduces the concept accuracy (row \#3 vs.\ \#1, Table \ref{sup:tab:ablation}). Whereas using the intervened cyclic loss $\mathcal{L}_{i_2}$ with the image reconstruction loss $\mathcal{L}_{r_2}$ improves the concept accuracy at the expense of image quality and steerability (row \#4 vs.\ \#1, Table \ref{sup:tab:ablation}). Finally, using both of the intervention losses achieves a better tradeoff between the three metrics (row \#5 vs.\ \#3, \#4, Table \ref{sup:tab:ablation}).

\noindent
\textbf{Scaling image resolution.} 
Based on Table \cvprcolor{2} and \cvprcolor{4} (main paper), our methods achieve good performance on PGAN and DDPM when the image resolution is scaled from 64$\times$64 to 256$\times$256. We further validate this with CelebA-HQ PGAN trained at 512$\times$512 in Table \ref{sup:tab:pgan_scaling_resolution}. While the steerability is relatively lower than at 256$\times$256, we still achieve fairly good steerability, \ie successful interventions with the same training time.

\begin{table}[]
\centering
\caption{\textbf{Sensitivity of CB-AE to number of concepts} for CelebA-HQ-StyleGAN2 using TIP-few-shot for pseudo-labels. Evaluation is done only on 8 shared concepts for a fair comparison. 
}
\vspace{-3mm}
\label{tab:scaling_num_conc}
\setlength{\tabcolsep}{3mm}
\resizebox{\linewidth}{!}{
\begin{tabular}{@{}lccc@{}}
\toprule
\multirow{2}{*}{\begin{tabular}[c]{@{}l@{}}Trained with \\ $M\!=\!$ TIP-fs-128\end{tabular}} &
  \multirow{2}{*}{\begin{tabular}[c]{@{}c@{}}Conc. Acc. \\ (\%)\end{tabular}} &
  \multicolumn{2}{c}{Steerability (\%)} \\ \cmidrule(l){3-4} 
            &       & CB-AE & CB-AE w/ opt-int \\ \midrule
8 concepts  & 76.08 & 21.51 & 38.73            \\
40 concepts & 75.75 & 22.17 & 39.94            \\ \bottomrule
\end{tabular}
}
\vspace{-3mm}
\end{table}

\noindent
\textbf{Sensitivity to intervention hyperparameters.} We analyze the sensitivity to optimization-based intervention hyperparameters in Fig.\ \ref{sup:fig:optint_sens}\cvprcolor{A}, \cvprcolor{B}. Since we used the iterative randomized fast gradient sign method \cite{wong2020fast}, the two hyperparameters involved are the number of iterations and the $\ell_\infty$-norm bound $\epsilon$ (maximum allowable perturbation). We find that as $\epsilon$ is increased, the steerability also increases but with a drop in image quality since the FID increases. Hence, we choose a small $\epsilon=0.1$ for most of our experiments such that we obtain a good tradeoff between image quality and steerability. Further, we observe that steerability and image quality remain similar when the number of iterations are reduced from 50 to 10 iterations. However, we use 50 iterations in our experiments to allow the optimization to converge for samples that are more difficult to intervene.

\noindent
\textbf{Sensitivity to CB-AE location.}
We vary the CB-AE location in CelebA-pretrained DCGAN and report the steerability and concept accuracy in Fig.\ \ref{sup:fig:optint_sens}\cvprcolor{C}.
We observed that CB-AE closer to generator output hurts steerability (decreased to 27.1\%) as modified latent has less influence on the output, but increased steerability up to 47.3\% near the middle. On the other hand, concept accuracy remains reasonable across all locations.

\begin{figure}
    \centering
    \includegraphics[width=\linewidth]{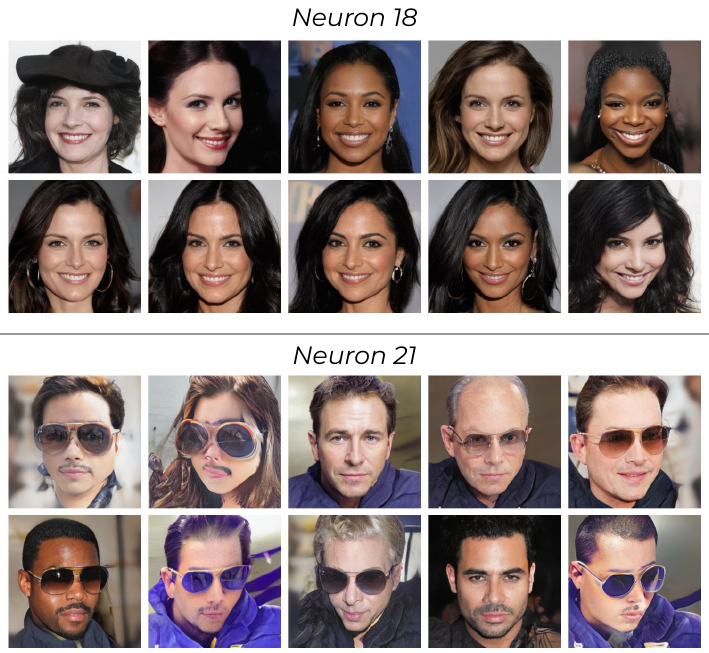}
    \caption{\revision{Top-10 images activating a particular neuron from the unsupervised concept embedding for CelebA-HQ StyleGAN2 CB-AE. We observe `earrings' (top) and `sunglasses' (bottom) concepts which were not present in the predefined concept set.}}
    \label{sup:fig:unsup_analysis}
\end{figure}

\noindent
\textbf{Unsupervised concept embedding analysis.}
For CB-AE trained with CelebA-HQ-pretrained StyleGAN2, we generated 5k images and collected top-10 images for each dimension in the unsupervised concept embedding being highly activated. 
Based on the common attributes in the top-10 images, we identified `sunglasses' and `earrings' (not in predefined concepts) as shown in Fig.\ \ref{sup:fig:unsup_analysis}.

\begin{figure*}
    \centering
    \includegraphics[width=\linewidth]{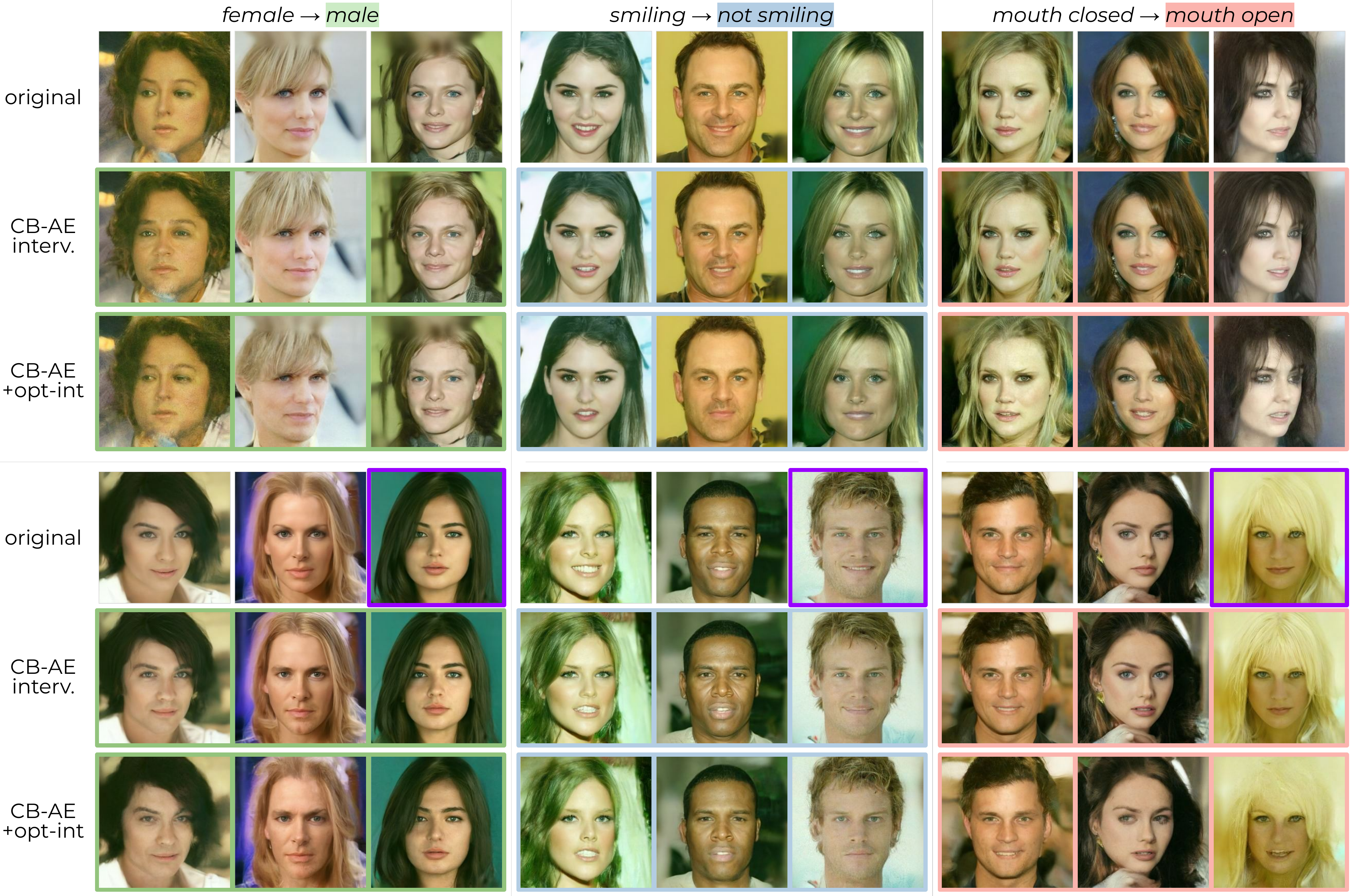}
    \caption{Concept intervention examples for CB-AE and CB-AE with optimization-based interventions (opt-int) for CelebA-HQ-pretrained DDPM. Some cases where either or both of our methods failed are highlighted in \textcolor{purple}{purple}.}
    \label{sup:fig:ddpm_examples}
\end{figure*}

\begin{figure*}
    \centering
    \includegraphics[width=0.9\linewidth]{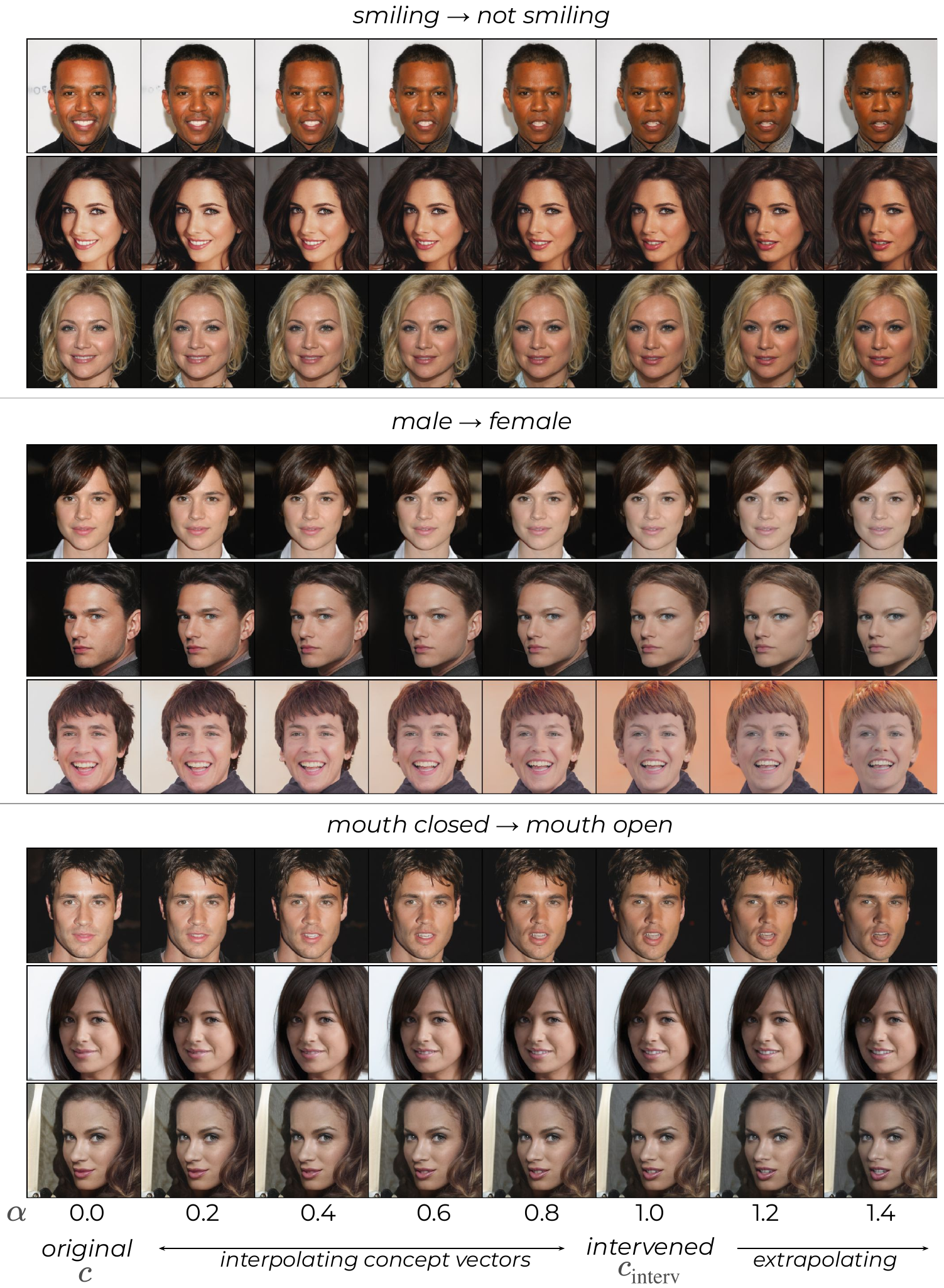}
    \caption{\textbf{Concept vector interpolation.} We interpolate between the concept vector $c$ from the CB-AE and the intervened concept vector $c_\text{intervened}$ for generating images, \ie $\hat{c}_\text{intervened}=(1-\alpha)c + \alpha c_\text{intervened}$ where $\alpha\in[0, 1]$. The interpolated vector $\hat{c}_\text{intervened}$ is passed through the CB-AE decoder $D$ and the remaining generator $g_2$ to obtain the displayed images. We also show examples with extrapolation for $\alpha=1.2, 1.4$.}
    \label{sup:fig:conc_interp_examples}
\end{figure*}

\noindent
\textbf{Qualitative evaluation.} In Fig.\ \ref{sup:fig:ddpm_examples}, we show concept intervention examples of our CB-AE and CB-AE with optimization-based interventions for a CelebA-HQ-pretrained DDPM diffusion model. Unlike with StyleGAN examples (Fig.\ \cvprcolor{4}, \cvprcolor{5}, main paper), we find that optimization-based interventions produce relatively lower quality images compared to CB-AE interventions. We also highlight some cases where either or both of our methods failed. In these cases, we find that some other concepts like hair style change marginally or the desired concept does not change enough.

\begin{table}[]
\centering
\caption{\textbf{Inference time analysis} for CB-AE with CelebA-HQ-pretrained StyleGAN2. Here, opt-int-$k$ indicates optimization-based interventions with $k$ iterations. Inference time (in milliseconds) is computed with batch size 64 on a single V100 GPU, repeated 1000 times for mean and standard deviation.}
\vspace{-3mm}
\label{sup:tab:efficiency}
\setlength{\tabcolsep}{10mm}
\resizebox{\linewidth}{!}{
\begin{tabular}{lc}
\toprule
 & Inference Time (ms) \\
\midrule
Base model & 170.02 $\pm$ 0.45 \\
CB-AE reconstr. & 170.70 $\pm$ 0.53 \\
CB-AE interv. & 170.27 $\pm$ 2.26 \\
CB-AE+opt-int-10 & 181.68 $\pm$ 2.98 \\
CB-AE+opt-int-50 & 226.01 $\pm$ 1.05 \\
\bottomrule
\end{tabular}
}
\vspace{-4mm}
\end{table}

\noindent
\textbf{Concept interpolation.}
To demonstrate that our training objectives incorporate meaningful knowledge in the CB-AE, we generate images using interpolation (and extrapolation) between predicted and intervened concept vectors, as shown in Fig.\ \ref{sup:fig:conc_interp_examples}. Concretely, for a randomly sampled noise vector $z$, we can compute the concept vector $c = E(g_1(z))$ using the CB-AE encoder $E$ and the first part of the generator $g_1$. Then, given a target concept, we compute an intervened concept vector $c_\text{intervened}$ as described in the CB-AE Objective 3 (Sec.\ \cvprcolor{3.1}, main paper). The interpolated concept vector can be computed as $\hat{c}_\text{intervened} = (1-\alpha)c + \alpha c_\text{intervened}$ where $\alpha\in[0, 1]$ (and extrapolation for $\alpha > 1$). Then, an image can be generated using the interpolated concept vector as $\hat{x}_\text{intervened} = g_2(D(\hat{c}_\text{intervened}))$ using the CB-AE decoder $D$ and $g_2$.

Overall, we observe that the CB-AE can produce smooth transitions in the image space from the original to intervened concept vectors as well as extrapolate further. However, in some of the extrapolation cases, we find changes in other concepts like hair color or skin color apart from the target concept. While it is generally undesirable for concept interventions, this can be a potential tool for dataset creators or generative model developers to identify potential biases or spurious correlations between concepts.

\begin{table}[]
\centering
\caption{\revision{\textbf{Trainable parameters analysis} for CB-AE and CC with CelebA-HQ-pretrained StyleGAN2 \wrt CBGM. Reduction indicates $\%$ reduction in trainable parameters compared to CBGM.}}
\vspace{-3mm}
\label{sup:tab:trainable_params}
\setlength{\tabcolsep}{4mm}
\resizebox{\linewidth}{!}{
\begin{tabular}{lcc}
\toprule
Method & Trainable Parameters & Reduction (\%) \\
\midrule
CBGM \cite{ismail2024concept} & 24.77M & - \\
CB-AE (\textit{Ours}) & 1.64M & 93.37 \\
CC (\textit{Ours}) & 0.79M & 96.77 \\
\bottomrule
\end{tabular}
}
\vspace{-4mm}
\end{table}


\subsection{Efficiency analysis}
\label{sup:subsec:efficiency_analysis}

In Table \ref{sup:tab:efficiency}, we compare the inference time of our methods with the base model. We compute the inference times for CB-AE trained with CelebA-HQ-pretrained StyleGAN2 using batch size 64 on a single V100 GPU. We repeat the inference 1000 times and report the mean and standard deviation, and find that using the CB-AE with the base model (in reconstruction mode, without interventions) and for concept interventions only causes a marginal increase in inference time. Given the number of iterations involved in optimization-based interventions, there is a relatively larger increase in inference time. However, as shown in Fig.\ \ref{sup:fig:optint_sens}, our method is effective even with 10 iterations, which adds only $\sim$11 milliseconds of inference time to that of the base model.

We also compare the number of trainable parameters in our CB-AE and CC compared to CBGM in Table \ref{sup:tab:trainable_params}.
Due to our efficient and novel autoencoder setup, we find 93.37\% and 96.77\% reduction in trainable parameters for StyleGAN2 compared to CBGM \cite{ismail2024concept}.

{
\small
\bibliographystyle{ieeenat_fullname}
\bibliography{ref}
}

\end{document}